\documentclass[preprint,12pt]{elsarticle}

\usepackage{amssymb}
\usepackage{amsmath}
\usepackage{amsthm}
\usepackage{booktabs}
\usepackage{hyperref}
\usepackage{pifont}  
\usepackage{algorithmic}
\usepackage[switch]{lineno}
\usepackage{subcaption}
\usepackage[ruled,linesnumbered]{algorithm2e}
\usepackage{makecell}
\usepackage{etoolbox}
\usepackage{multirow}

\newcommand{\cmark}{\checkmark}
\newcommand{\xmark}{\ding{55}}

\journal{Knowledge-Based Systems}

\begin{document}

\begin{frontmatter}



\title{PromptAL: Sample-Aware Dynamic Soft Prompts for Few-Shot Active Learning}




\author[1,3]{Hui Xiang}
\ead{xianghui0325@bupt.edu.cn}
\author[1,2]{Jinqiao Shi\corref{corresponding}}
\ead{shijinqiao@bupt.edu.cn}
\author[3]{Ting Zhang}
\ead{douglasleft@gmail.com}

\author[1,2]{Xiaojie Zhao}
\ead{xiaojiezhao@bupt.edu.cn}

\author[3,5]{Yong Liu}
\ead{liuyong03@qianxin.com}

\author[3,4]{Yong Ma\corref{corresponding}}
\ead{znsoft@163.com}

\cortext[corresponding]{Corresponding author.}
\affiliation[1]{organization={School of Cyberspace Security, Beijing University of Posts and Telecommunications}, 
            postcode={100876},
            city={Beijing},
            country={China}}
\affiliation[2]{organization={Key Laboratory of Trustworthy Distributed Computing and Service (BUPT), Ministry of Education}, 
            postcode={100876}, 
            city={Beijing},
            country={China}}

\affiliation[3]{organization={QI-ANXIN  Group},
            	addressline={QAX Security Center,No.1,26 of Xizhimenwai South Road,Xicheng District}, 
            	city={Beijing},
            	postcode={100000}, 
            	state={Beijing},
            	country={China}}
\affiliation[4]{organization={RapidAI Research},
            	addressline={Haidian District}, 
            	city={Beijing},
            	postcode={100085}, 
            	state={Beijing},
            	country={China}}
\affiliation[5]{organization={ZGC Lab},
            	addressline={Haidian District}, 
            	city={Beijing},
            	postcode={100085}, 
            	state={Beijing},
            	country={China}}
\begin{abstract}
Active learning (AL) aims to optimize model training and reduce annotation costs by selecting the most informative samples for labeling.
Typically, AL methods rely on the empirical distribution of labeled data to define the decision boundary and perform uncertainty or diversity estimation, subsequently identifying potential high-quality samples.
In few-shot scenarios, the empirical distribution often diverges significantly from the target distribution, causing the decision boundary to shift away from its optimal position.
However, existing methods overlook the role of unlabeled samples in enhancing the empirical distribution to better align with the target distribution, resulting in a suboptimal decision boundary and the selection of samples that inadequately represent the target distribution.
To address this, we propose a hybrid AL framework, termed \textbf{PromptAL} (Sample-Aware Dynamic Soft \textbf{Prompts} for Few-Shot \textbf{A}ctive \textbf{L}earning).
This framework accounts for the contribution of each unlabeled data point in aligning the current empirical distribution with the target distribution, thereby optimizing the decision boundary.
Specifically, PromptAL first leverages unlabeled data to construct sample-aware dynamic soft prompts that adjust the model's predictive distribution and decision boundary.
Subsequently, based on the adjusted decision boundary, it integrates uncertainty estimation with both global and local diversity to select high-quality samples that more accurately represent the target distribution.
Experimental results on six in-domain and three out-of-domain datasets show that PromptAL achieves superior performance over nine baselines.
Our codebase is openly accessible.
\end{abstract}

\begin{graphicalabstract}
\end{graphicalabstract}

\begin{highlights}

\item Sample-aware soft prompts are introduced into active learning for dynamically adjusting the probability distribution and optimizing the decision boundary

\item Innovatively represent uncertainty and diversity scores by leveraging knowledge feature space

\item A hybrid query strategy that effectively balances uncertainty and diversity enables the selection of more representative samples.

\end{highlights}


\end{frontmatter}



\section{Introduction}
Predominant supervised learning approaches in natural language processing (NLP) heavily rely on vast amounts of annotated data.
In practice, annotating such data necessitates substantial human effort and time resources.
Active learning (AL) is an extensively researched domain aimed at reducing annotation costs by selectively querying the most informative samples for labeling and model training \cite{zhangSurveyActiveLearning2022, least-confidence, ashDeepBatchActive2019, cal}.
\begin{figure}[t]
  \centering
\includegraphics[width= 0.7\textwidth]{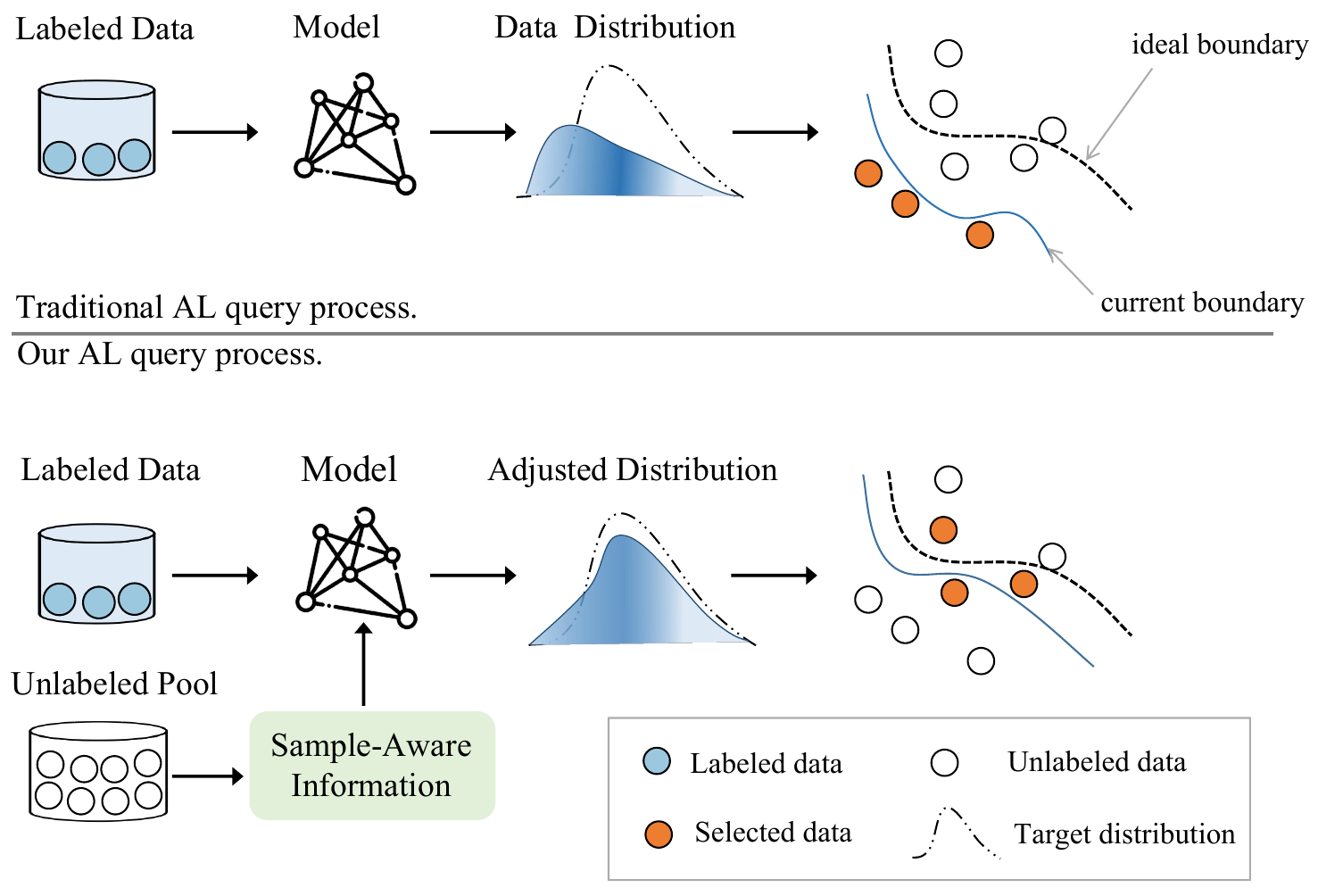}
  \caption{ Motivation behind PromptAL.} 
  \label{figure1}
\end{figure}
AL methodologies typically operate under the assumption that the model learns the data distribution from an initial set of labeled samples, leveraging this distribution to estimate uncertainty or diversity for selecting high-quality samples for training~\cite{least-confidence, cal, entropy}.
However, in few-shot scenarios, the limited availability of labeled samples results in a significant bias between the empirical and target distributions \cite{10.1145/3582688}.
This bias causes the decision boundary used for sample selection to shift away from its optimal position, leading to significant deviations in sample evaluation.
Therefore, certain samples deemed as high quality are not genuinely high quality and may have negligible or even adverse effects on fitting the target distribution, potentially impeding the improvement of the model performance.
Consequently, AL algorithms encounter substantial challenges in few-shot settings, thereby necessitating the development of more advanced strategies.
To address these challenges, it is crucial to explore additional sources of information beyond labeled samples.
We observed that leveraging information from unlabeled data significantly enhances the alignment of the model with the underlying target distribution, particularly in few-shot learning scenarios.
Building on this insight, we concentrated on extracting sample-aware information from unlabeled data to improve distribution fitting.
This information is utilized to refine the predictive distribution of the model, thereby adjusting the decision boundary.
Consequently, the query strategy selects samples that more accurately represent the target distribution, effectively boosting the model performance.
The motivation for our approach is illustrated in Fig.\ref{figure1}.

To achieve this, we introduce a sample-aware dynamic soft prompt that modifies the predictive distribution based on unlabeled samples.
Furthermore, we propose a novel AL framework, \textbf{PromptAL} (Sample-Aware Dynamic Soft \textbf{Prompts} for Few-Shot \textbf{A}ctive \textbf{L}earning), which efficiently queries samples essential for accelerating convergence toward the target distribution in few-shot learning scenarios.

Specifically, PromptAL is a hybrid AL framework that strategically integrates both uncertainty and diversity principles.
For uncertainty estimation, PromptAL calculates entropy scores using the predictive distribution derived through dynamic soft prompts.
However, an exclusive focus on uncertainty can result in redundant query sets, thereby diminishing the generalization capability of the model \cite{dor2020active}.
To address this issue, PromptAL integrates both global and local diversity into the sample selection process.
Global diversity is achieved by clustering in the knowledge feature space derived from dynamic soft prompts, ensuring that the selected samples represent the feature distribution of the unlabeled dataset.
Local diversity ensures that the selected samples exhibit distinct characteristics compared to the labeled data, thereby reducing redundancy in the selection process.
{Finally, PromptAL selects samples with the highest joint scores, which combine uncertainty and local diversity through a weighted sum.}
In summary, the contributions of this study are listed as follows:
\begin{itemize}
    \item 
    We introduce PromptAL, an innovative hybrid AL framework that incorporates sample-aware dynamic soft prompts as a core component, facilitating the selection of samples that significantly contribute to the convergence of the model toward the target distribution.
    Notably, to the best of our knowledge, this study is the first to integrate soft prompt mechanisms into the AL paradigm, opening new avenues for further exploration in the field.
    \item Extensive experiments demonstrate that PromptAL outperforms nine baseline methods across six in-domain and three out-of-domain (OOD) tasks, while ablation studies further validate the effectiveness of each component within PromptAL.
    \item We conduct a comprehensive analysis of the selected samples, showing that PromptAL achieves a superior balance among category distribution, representativeness, uncertainty, and diversity compared to other methods.
\end{itemize}
The source code for this study is available on GitHub at \href{https://github.com/xiang-hui744/PromptAL}{https://github.com/PromptAL}.

\section{Related Work}
\subsection{Active Learning}
The classic AL query strategies generally adopt either uncertainty-based or diversity-based approaches, commonly described as the “two faces of AL"~\cite{dasgupta2011two}.
Hybrid query strategies that integrate uncertainty and diversity have recently emerged.
\subsubsection{Uncertainty Sampling}
Uncertainty sampling is a prevalent query strategy targeting the most uncertain data points as determined by model predictions.
The least-confidence (LC) method~\cite{least-confidence} exemplifies an uncertainty-based approach by selecting samples with the lowest maximum probability for annotation.
Similarly, entropy~\cite{entropy} is another established technique that chooses data points where the model exhibits the highest predictive entropy.
In few-shot scenarios, the scarcity of training data leads models to develop a pronounced bias toward the target distribution, resulting in inaccurate probability estimates.
Consequently, uncertainty scores become unreliable, and samples with high uncertainty do not necessarily provide substantial benefits to model learning.
\subsubsection{Diversity Sampling}
In contrast, diversity sampling operates on the principle that a labeled, representative subset of data can effectively capture the key characteristics of the entire dataset \cite{tang2002active,maekawa2022low}.
For example, BERT-KM~\cite{bertkm} utilizes the output embeddings from a fine-tuned BERT model as a feature space and applies k-means clustering to select sample points from each cluster.
%
Additionally, Core-set~\cite{Core} is a diversity-based query strategy applied to convolutional neural networks that selects a subset of points such that a model trained on this subset remains competitive when evaluated on the remaining data points.
\subsubsection{Hybrid Sampling}
Hybrid strategies take both uncertainty and diversity into account \cite{shi2021diversity,zhdanov2019diverse}.
%
Batch AL by Diverse Gradient Embeddings (BADGE)~\cite{ashDeepBatchActive2019} is a leading method that leverages the magnitude of parameter gradients in the final output layer as an objective measure of uncertainty, selecting a batch of instances with gradients spanning diverse directions to ensure variability.
Recently, Contrastive AL (CAL)~\cite{cal} has emerged as a hybrid query strategy that identifies contrastive samples---data points that are similar in feature space but elicit maximally different predictive likelihoods from the model.

Additionally, several hybrid query strategies have been developed to address cold-start or few-shot scenarios \cite{yuanColdstartActiveLearning2020}.
For example, AL by Processing Surprisal (ALPS) employs the masked language modeling (MLM) loss as a proxy for classification uncertainty and subsequently clusters surprisal embeddings to minimize labeling costs.
Recent approaches also combine manual prompts with AL to leverage the model's task-specific knowledge for sample selection in cold-start or few-shot settings.
MEAL~\cite{koksalMEALStableActive2023} integrates multiple hard prompts with AL to enhance the stability of prompt tuning.
Patron~\cite{yuColdStartDataSelection2023} introduces an uncertainty propagation method based on hard prompts to assess the importance of data points.
However, hard prompts rely on fixed artificial task settings, whereas standard soft prompts remain unchanged after training.
Therefore, these two prompts are static during an inference process, and neither of them fully addresses the impact of unlabeled samples on the advancement of the model toward the target distribution.
\subsection{Prompt Learning}
Recently, the effectiveness of pre-trained language models (PLMs)~\cite{devlin2018bert,liu2019roberta} has been proven in NLP tasks, as PLMs can grasp linguistic~\cite{jawahar2019does}, semantic~\cite{yenicelik2020does}, syntactic~\cite{hewitt2019structural}, and world knowledge~\cite{petroni2019language} from extensive corpora.
However, in low-resource settings, the substantial disparity between pre-training tasks (e.g., MLM) and fine-tuning tasks (e.g., classification and regression) impedes rapid adaptation of the model to downstream tasks.
Fortunately, prompt learning is an emerging machine learning paradigm that has shown remarkable performance in few-shot and zero-shot scenarios \cite{liuPretrainPromptPredict2023a}.
This success is mainly due to its ability to bridge the gap between pre-training and downstream task phases, effectively utilizing the knowledge embedded in PLMs.
A prompt typically consists of two essential components: a prompt template $\mathcal{T}$, which directs the model to convert the task into a cloze-style format, and a verbalizer $\mathcal{V}$, which translates the predicted label words back to their corresponding labels.
Prompt templates are classified into hard prompts and soft prompts based on whether they use discrete or continuous representations.
\subsubsection{Hard Prompt}
A hard prompt is a carefully constructed prompt composed of a set of natural language words \cite{schickExploitingClozeQuestionsFewShot2021c}.
For example, in the movie review classification task, we use the prompt template $\mathcal{T}$(\textit{x}) = \underline{[\textit{x}]. It was \texttt{<MASK>}.} and the verbalizer $\mathcal{V}$ = \{1: \textit{great}, 0: \textit{bad}\}, where [\textit{x}] denotes the original text, and \textit{great} and \textit{bad} serve as the label words of the verbalizer.
PLMs classify [\textit{x}] as either category 1 or 0 by predicting whether the word at the \texttt{<MASK>} position is \textit{great} or \textit{bad}.
\subsubsection{Soft Prompt}
Soft prompts utilize gradient descent optimization to identify optimal vectors within the model’s continuous embedding space, functioning as prompts instead of fixed discrete tokens \cite{gu-etal-2022-ppt,liuGPTUnderstandsToo2024}.
Unlike hard prompts, soft prompts adapt to various tasks.
%
For example, for the above task, the soft prompt template can be defined as: $\mathcal{T}$(\textit{x}) = \underline{ [\textit{$v_{1}$}][\textit{$v_{2}$}]...[\textit{$v_{i}$}] [\textit{x}] \texttt{<MASK>}.} where [\textit{$v_{i}$}] represents the trainable pseudo-token vector.
Beyond task-specific soft prompts, several studies have identified instance-specific soft prompts \cite{wuIDPGInstanceDependentPrompt2022,jainPromptTuningStrikes2024a}.

In contrast to these approaches, our method employs sample-aware dynamic soft prompts to adaptively modulate the predictive distribution.
By integrating both uncertainty and diversity, we strategically select samples that enhance the model's alignment with the target distribution.

\section{Active Learning Loop}
This section delineates the AL loop process.
Formally, we address the classification task with $C$ categories using a model $\mathcal{M}$, where $Y =\left \{ \mathit{y_{i} } \right \}_{i=1}^{C} $ denotes the set of categories.
Assume the existence of a substantial unlabeled pool $D_{pool} =\left \{ \mathit{x_{i} } \right \}_{i=1}^{N} $ of N samples and an initial training set $D_{train}$.
Given a total annotation budget of $B$, each AL iteration involves selecting a subset $Q$ from \( D_{\text{pool}} \) for manual annotation and incorporating $Q$ into the training set.
Subsequently, $\mathcal{M}$ is fine-tuned on the augmented training data.
After $t = |B|/|Q|$ iterations, the AL process concludes, resulting in a trained model and a labeled dataset.
The primary objective of AL is to enhance the model's accuracy while minimizing dependence on labeled data.
The specific procedure is detailed in Algorithm \ref{ALalgorithm}.
\begin{algorithm}
\caption{Overall process of the AL loop }\label{ALalgorithm}
\KwIn{query set $Q=\emptyset$, model $\mathcal{{M}}_0$, train set $D_{train}$, unlabeled pool $D_{pool}$,  iterations $t$, AL query strategy }
\For{$i$ is 1,2,... $t$}
{$Q\gets$ apply AL query strategy on $\mathcal{M}_{i-1}$ in $D_{pool}$ \; 

$D_{pool}=D_{pool} \setminus Q $ \;

human annotate $Q$ \; 
 
$D_{train}=D_{train} \cup {Q}$ \;

 $\mathcal{{M}}_i \gets$ tune $\mathcal{{M}}_0$ on $D_{train}$\; 
}
\Return{$\mathcal{{M}}_t,$ $D_{train}$}
\end{algorithm}

\section{Methodology}
\begin{figure*}
    \centering
\includegraphics[width=1.1\textwidth]{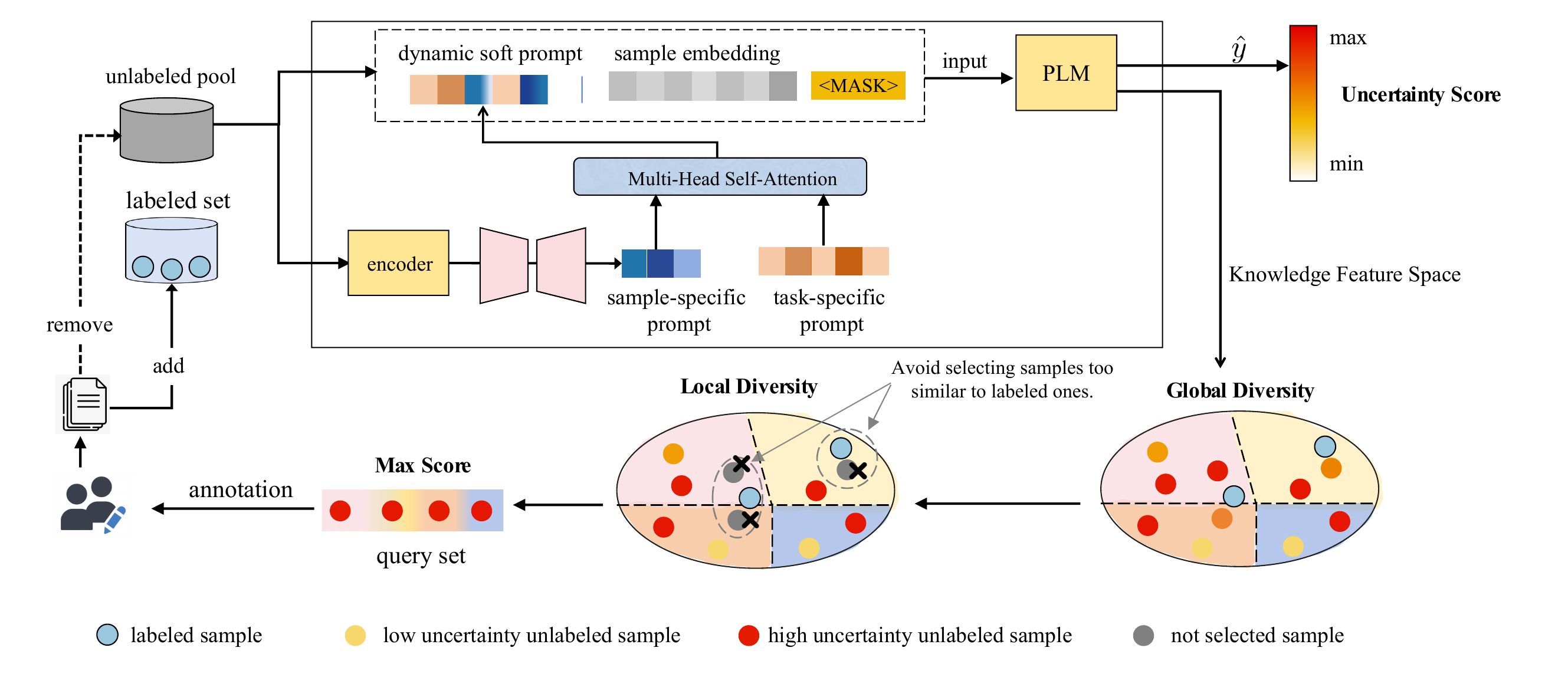}
 \caption{{Principle of PromptAL Framework. The process consists of four main steps: (1) generating sample-aware dynamic soft prompts to adjust the predictive distribution of PLMs for unlabeled samples; (2) estimating uncertainty scores based on predictive distributions; (3) performing global clustering and computing local diversity scores for each sample within each cluster; (4) samples with the highest joint scores—computed as a weighted combination of uncertainty and local diversity scores—are selected to form the query set.}}
    \label{PromptAL}
\end{figure*}

In this section, we introduce the foundational principle of PromptAL. PromptAL is a framework that integrates AL with soft prompts, comprising four stages, as illustrated in Fig.\ref{PromptAL}.
(1) Sample-aware dynamic soft prompts adjust the predictive distribution of PLMs for unlabeled samples; (2) Compute the uncertainty score of each sample based on its predictive distribution to identify the most informative uncertain samples;
(3) Cluster the samples into globally diverse groups within the knowledge feature space to enhance global diversity, and evaluate the similarity of each sample to labeled data to measure local diversity;
(4) Within each globally diverse cluster, select the sample with the maximum joint score combining uncertainty and local diversity as the query set, ensuring a balance between uncertainty and diversity in the selection process.
The query set is then annotated by domain experts and incorporated into the training set for the subsequent AL iteration.
\subsection{Sample-aware Dynamic Soft Prompt}
\label{3.1}
PromptAL integrates task-specific and sample-specific soft prompts via a multi-head self-attention mechanism, thereby generating sample-aware dynamic soft prompts.
This mechanism enables the dynamic soft prompts to adapt to different input samples.


For sample $x$, we define the dynamic soft prompt as \( \mathbf{P}(x) \), which is generated by integrating the task-specific prompt \( \mathbf{T} \in \mathbb{R}^{m \times d} \) and the sample-specific prompt \( \mathbf{S}(x) \in \mathbb{R}^{n \times d} \). \( \mathbf{T} \) is a learnable parameter matrix shared across all samples, while \( \mathbf{S}(x) \) is unique to each sample. \( m \) and \( n \) denote the number of vectors, and \( d \) is the vector dimension.

To generate \( \mathbf{S}(x) \), we design a sample prompt generator \( f \), which combines the encoder \( E \) and a two-layer multi-layer perceptron (MLP). Given a sample \( x \), the encoder \( E \) first encodes \( x \) into \( E(x) \in \mathbb{R}^d \). Subsequently, the first layer of the MLP maps \( E(x) \) from \( \mathbb{R}^d \) to  \( \mathbb{R}^{l} \), and the second layer maps it to \(\mathbf{S}(x) \in  \mathbb{R}^{n \times d} \).

\( \mathbf{P}(x) \) $\in$ \( \mathbb{R}^{(m+n) \times d} \) is the fusion of \( \mathbf{T} \) and \( \mathbf{S}(x) \) through a multi-head self-attention mechanism:
\begin{equation}
\mathbf{P}(x) = \text{MultiHead} \left( \text{concat} \left[ \mathbf{T}, \mathbf{S}(x) \right] \right) 
\label{px}
\end{equation}

In this mechanism, the \( i \)-th attention head generates a prompt vector for $x$, denoted as $\mathbf{P}_i(x)$:
\begin{equation}
\mathbf{P}_i(x) = \text{softmax} \left( \frac{\mathbf{W}_Q^{(i)} \mathbf{P^*}(x) (\mathbf{W}_K^{(i)} \mathbf{P^*}(x))^\top}{\sqrt{d}} \right) \mathbf{W}_V^{(i)} \mathbf{P^*}(x)
\end{equation} where \( \mathbf{P^*}(x) \) is \( \text{concat} \left[ \mathbf{T}, \mathbf{S}(x) \right] \) which belongs to \( \mathbb{R}^{(m+n) \times d} \). 
\( \mathbf{W}_Q^{(i)} \), \( \mathbf{W}_K^{(i)} \), and \( \mathbf{W}_V^{(i)} \) represent the learnable weight matrices corresponding to the query, key, and value, respectively, for the \( i \)-th attention head.
The outputs from all attention heads are concatenated and subsequently subjected to a linear transformation via the weight matrix, yielding the final dynamic soft prompt \( \mathbf{P}(x) \).

After obtaining \( \mathbf{P}(x) \), we formulate the input prompt for the pretrained model $\mathcal{M}$, which utilizes MLM capabilities, as follows:
\begin{equation}
    \mathcal{T}(x)  = \text{concat}[\mathbf{P}(x),  \boldsymbol{\phi}(x), \boldsymbol{\phi}(\texttt{<MASK>})]
     \label{Tx}
\end{equation}
where \( \boldsymbol\phi(\cdot) \) denotes the token embedding function and \texttt{<MASK>} refers to the mask token of model $\mathcal{M}$.
%

The probability distribution for \( x \in \text{label } y \) is as follows:
\begin{align}
p(y|x) &= p( \texttt{<MASK>} = \mathcal{V}(y) |  \mathcal{T}(x) ) 
\end{align}
%
%
where \(\mathcal{V}(y)\) denotes the label word corresponding to label \( y \) in the verbalizer $\mathcal{V}$.
\subsection{Uncertainty Estimation}
PromptAL employs calibrated probability distributions to quantify the uncertainty of samples.
Owing to imbalanced pre-training corpora, certain words are more likely to be predicted than others \cite{zhao2021calibrate}.
For instance, a model pre-trained predominantly on positive data tends to predict \textit{positive} over \textit{negative}, regardless of the semantics of the input.
To mitigate this, following KPT \cite{huKnowledgeablePrompttuningIncorporating2022c}, we employ the \textit{contextualized prior} to calibrate the probabilities.
We first construct a support set $\mathcal{R}$ by selecting the top $k$ samples with the highest $p(y_{i} | x)$ for each label $y_{i} \in Y$.
The \textit{contextualized prior} is then estimated as
\begin{equation}
    p(\nu) \approx \frac{1}{|\mathcal{R}|} \sum_{x \in \mathcal{R}} p
    (\texttt{<MASK>} = \nu |  \mathcal{T}(x))
\end{equation}
where $\nu$ denotes the word from the label word set of $\mathcal{V}$.

The calibrated probability distribution is as follows:
\begin{equation}
    \hat{y}_i = \left( \frac{p(y_i | x)}{p(\nu(y_i))} \right)/ \left( \sum_{j=1}^{C} \frac{p(y_j | x)}{p(\nu(y_j))} \right)
    \label{calibrated}
\end{equation}
where $\hat{y_i}$ represents the calibrated probability that $x$ belongs to category $y_i$.
Finally, the uncertainty score of $x$ is calculated using the entropy formula~\cite{lewis1995sequential}.
\begin{equation}
    U(x) = - \sum_{i=1}^{C} \hat{y}_i \log \hat{y}_i.
    \label{u(x)}
\end{equation}
\subsection{Diversity Estimation}
\label{3.2}
Diversity estimation is assessed from two angles: global diversity and local diversity.
\subsubsection{Global Diversity with Knowledge Features}
{PromptAL clusters within a knowledge feature space to select samples that capture the overall feature distribution of the unlabeled data, thereby ensuring global diversity.}
Inspired by PromptBERT~\cite{jiangPromptBERTImprovingBERT2022c}, we adopt the vector of the mask token from the last hidden layer, denoted as $\boldsymbol{\mathit{h}}_{\texttt{<MASK>}}$, as the knowledge feature to construct the knowledge feature space.
The knowledge feature at the {\texttt{<MASK>}} position, activated by dynamic soft prompts, encapsulates the comprehensive knowledge from model $\mathcal{M}$ to sample $x$ in the current task, surpassing mere semantic information \cite{liuPretrainPromptPredict2023a}.
We employ K-means++ clustering \cite{kmeans++} on $\boldsymbol{\mathit{h}}_{x}$ (the $\boldsymbol{\mathit{h}}_{\texttt{<MASK>}}$ for sample $x$) within the unlabeled dataset, dividing it into $|Q|$ clusters.

\subsubsection{Local Diversity}
PromptAL introduces local diversity to prevent selecting samples that are excessively similar to those in the training set.
Specifically, PromptAL utilizes K-Nearest Neighbors (KNN)~\cite{knn} to identify the \( k' \) closest training samples relative to $x$.
The mean distance of these \( k' \) samples is calculated as the local diversity score $D(x)$.
\begin{equation}
    D(x) = \frac{1}{k'} \sum_{i=1}^{k'} \| \boldsymbol{\mathit{h}}_{x} - \boldsymbol{\mathit{h}}_{x_i} \|_2
    \label{d(x)}
\end{equation}
where the distance is the Euclidean distance.
\subsection{Joint Score}

\label{3.3}
PromptAL integrates the uncertainty score with the local diversity score, weighted by \( \lambda \), to derive the final joint score \( S(x) \).
The parameter \( \lambda \in [0, 1] \) controls the relative weighting of the two scores, where higher values of \( \lambda \) place more emphasis on the uncertainty score.

\begin{equation}
    S(x) = \lambda \cdot U(x) + (1 - \lambda) \cdot D(x)
    \label{s(x)}
\end{equation}
Finally, PromptAL queries the samples with the highest value of joint score $S(x)$ from each cluster formed through global diversity to form the query set $Q$.
The PromptAL process is outlined in Algorithm~\ref{single}.
\begin{algorithm}[tb]
\small
\caption{Single iteration of PromptAL}\label{single}
\KwIn{query set $Q=\emptyset$, train set $D_{train}$, unlabeled pool $D_{pool}$, pre-trained model $\mathcal{M}$ equipped with MLM capability, dynamic soft prompt $\mathbf{P}(.)$, weight $\lambda$, acquisition number $b$ }
\For{$x$ in $D_{pool}$}
{$\mathbf{P}(x)\gets x  $ by equation \ref{px}\; 
$\mathcal{T}(x) \gets \mathbf{P}(x)  $ by equation \ref{Tx}\ ,{where
\( \mathbf{P}(x) \) denotes the dynamic soft prompt, and $\mathcal{T}(x)$ is the input sequence obtained by concatenating \( \mathbf{P}(x) \), the sample \( x \), and a \texttt{<MASK>} token;
}
\\
Feed $\mathcal{T}(x)$ into \(\mathcal{M}\) for prediction;
\\
compute $p(y|x)$ and calibration by equation \ref{calibrated} \;
compute uncertainty score $U(x)$ by equation \ref{u(x)}\;
compute local diversity score $D(x)$ by equation \ref{d(x)}\;
compute joint score $S(x)$ by equation \ref{s(x)}\;
}
knowledge feature space $\mathcal{D} =\left \{\boldsymbol{\mathit{h}}_{x} |x \in D_{pool}\right \}$ \;
$\mathcal{Q}\gets $k-MEANS++ cluster batch set of $\mathcal{D}$\;
$Q=\left \{ \underset{x\in  \mathcal{Q }_{i}}{argmax S(x)} |\mathcal{Q }_{i} \in\mathcal{Q} \right \} $\;
\KwOut{$Q$, $\text{ }|Q|=b$}
\end{algorithm}

\section{Experiments and Results}
The experimental settings are presented in Section \ref{setup}, and seven AL baselines are introduced in Section \ref{Baselines}.
We conduct five distinct experiments to demonstrate the superiority of PromptAL.
(1) The primary experiments (§\ref{Main Results}) show that PromptAL achieves higher accuracy than other baselines in few-shot scenarios.
(2) OOD experiments (§\ref{ood_Analysis}) assess the generalization capabilities of PromptAL.
(3) Comparative ablation studies (§\ref{Ablation Study}) evaluate the effectiveness of each PromptAL component.
(4) Hyperparameter analysis (§\ref{Hyperparameter Analysis}) explores the impact of hyperparameters to determine optimal values.
(5) A distribution alignment experiment (§\ref{Distribution Alignment}) validates the fundamental motivation behind PromptAL.

\subsection{Experiment Setup}
\label{setup}
\subsubsection{Datasets}
The experiments are performed on six datasets for in-domain tasks: IMDB \cite{imdb}, AGNews \cite{agnews}, DBpedia \cite{lehmann2015dbpedia}, Yelp-full \cite{yelpfull}, TREC \cite{trec}, and Yahoo!Answers \cite{agnews}.
For the Yahoo and DBpedia datasets, we randomly sampled 10k samples from each category in the full training set to include in the initial unlabeled pool.
For other datasets, the original training set serves as the initial unlabeled pool.
The statistical details of these datasets are presented in Table \ref{tab:dataset}.

For the OOD evaluation task, we employ three supplementary datasets to represent the OOD distribution for the IMDB dataset: SST-2 \cite{socher2013recursive}, IMDB Contrast Set (IMDB-Contrast) \cite{Contrast}, and IMDB Counterfactually Augmented Dataset (IMDB-Counter) \cite{Counterfactually}.
The SST-2 dataset \cite{socher2013recursive} is a concise collection of movie reviews for sentiment classification, which was evaluated using its validation set.
The IMDB Contrast Set \cite{Contrast} comprises test instances where ground-truth labels have been manually modified in a minor yet semantically significant manner by NLP researchers.
The IMDB Counterfactually Augmented Dataset \cite{Counterfactually} includes samples from the original IMDB dataset that have been slightly altered to invert sentiment labels while preserving most of the original content.
%
\begin{table}
    \centering
     \caption{Dataset statistic information.}
      \resizebox{0.8\textwidth}{!}{
    \begin{tabular}{cccccc}
        \toprule
        Dataset  &Domain & Class & Train  & Dev & Test \\
        \midrule
        IMDB     &Movie Review& 2  & 17.5k & 0.75k  & 25k       \\
        AGNews   &News Topic& 4     & 110k & 10k & 7.6k  \\
        DBpedia   &Wikipedia Text& 14    & 140k & 20k & 70k \\
        Yelp-full &Restaurant Review  & 5     & 35k  & 5k  & 10k \\
        TREC  &Web Text& 6     & 5k   & 0.4k    & 0.5k    \\
        Yahoo!Answers &Web QA& 10    & 100k & 20k & 60k \\
    
        \bottomrule
    \end{tabular}
    }
   
    \label{tab:dataset}
\end{table}

\begin{table*}[!ht]
    \centering
     \caption{Manual prompt and label words of the verbalizer.}
    \label{prompt}
    \resizebox{1.0\textwidth}{!}{
    \begin{tabular}{cccc}
        \toprule
        Dataset & Manual Prompt & Label words \\
        \midrule
        IMDB    & [\textit{x}]. It is \texttt{<MASK>}. & terrible, great \\
        AGNews    & \texttt{<MASK>} News:[\textit{x}] & World, Sports, Business, Tech \\
        DBpedia  & [\textit{x}]. It is a \texttt{<MASK>}. & \makecell[c]{Company, School, Artist, Athlete, Politics,\\ Transportation, Building, Mountain, Village, Animal,\\ Plant, Album, Film, Book} \\
        Yelp-full & [\textit{x}]. It is \texttt{<MASK>}. & terrible, bad, okay, good, great \\
        TREC   & [\textit{x}]. It is \texttt{<MASK>}. & Expression, Entity, Description, Human, Location, Number \\
        Yahoo!Answers & [Category: \texttt{<MASK>}][\textit{x}] & \makecell[c]{Society, Science, Health, Education, Computer, \\Sports, Business, Entertainment, Relationship, Politics} \\
    
        \bottomrule
    \end{tabular}
    }
   
\end{table*}

\subsubsection{Manual Prompt \& Verbalizer}
For baselines employing hard prompts, the manual prompt templates and label word configurations from the verbalizer proposed by \cite{yuColdStartDataSelection2023} were utilized, as illustrated in Table \ref{prompt}.

\subsubsection{Implementation Details}
The experimental parameters are detailed in Table \ref{tab:hyperparameters}.
We used RoBERTa-base \cite{liu2019roberta} as $\mathcal{M}$, sourced from the Huggingface Transformers library~\cite{hugging}.
To simulate a few-shot learning scenario, the initial training set consisted of 32 samples, followed by 10 AL iterations, each selecting 32 samples for annotation.
The AdamW optimizer with linear decay \cite{adamw} was applied, with a learning rate of 2e-5 and a training batch size of 8.
For the dynamic soft prompt, the vector number $n$ of the task-specific prompt is 4, the vector number $m$ of the sample-specific prompt is 1, and the vector dimension $d$ is 768.
Encoder $E$ denotes the model $\mathcal{M}$ obtained from the preceding AL iteration, which transforms $x$ into the \texttt{[CLS]} representation from the last hidden layer.
For the sample prompt generator $f$, the dimensionality of $\mathbb{R}^l$ is set to 256.
The self-attention mechanism comprises four heads with a hidden size of 768.
We utilize $k$ samples to construct the support set $R$ and the $k'$ = 10 nearest labeled samples to compute $D(x)$ as defined in Eq.\ref{d(x)}.
The weight parameter $\lambda$ is set to 0.9.
All experiments are conducted on an RTX A100 GPU and are repeated five times using different random seeds.
\begin{table}[ht]
\centering
\caption{Experimental setup.}
\begin{tabular}{ll}
 \toprule
\textbf{Item}                      & \textbf{Value}                                    \\ \hline
Pre-trained language model& RoBERTa-base    \\ 
Size of the initial training set               & 32 samples                                         \\ 
 AL iteration number $t$                   & 10                                       \\ 
acquisition number $b$                           & 32 samples per query  \\ 
Optimizer                                    & AdamW   \\ 
Learning rate                               & 2e-5   \\ 
Training batch size                         & 8  \\ 
vector number of task-specific prompt \( n \)   & 4                                                  \\ 
vector number of sample-specific prompt \( m \)   & 1                                                  \\ 
Vector dimension of dynamic soft prompt \( d \)   & 768                                                \\ 
Encoder \( E \) & \( \mathcal{M} \) from previous iteration  \\ 
Dimension of $\mathbb{R}^l$ in sample prompt generator $f$ & 256                                                \\ 
Number of self-attention heads              & 4                                                  \\ 
Hidden dimension of self-attention head                         & 768                                                \\ 
 The \( k \) samples used to construct support set                 & 100                                                \\ 
The \( k'\) closest labeled samples used to compute $D(x)$           & 10                                                 \\ 
Weight \( \lambda \) in joint scores                     & 0.9                                                \\ 
Hardware                                    & RTX A100 GPU                                       \\ 
Seed number                      & 5 (with random seeds)                             \\ 
\bottomrule
\end{tabular}
\label{tab:hyperparameters}
\end{table}
\subsection{Baselines}
\label{Baselines}
We compare PromptAL with { nine} query strategies, including uncertainty-based strategies: Entropy \cite{lewis1995sequential}, LC \cite{least-confidence}, and {BALD \cite{BALD}}; diversity-based strategies: BERT-KM \cite{bertkm} and { Core-set \cite{sener2018active}}; hybrid strategies: Patron \cite{yuColdStartDataSelection2023}, CAL \cite{cal}, and BADGE \cite{ashDeepBatchActive2019}; and the random sampling method.
\\
$\circ$ \textbf{Entropy} The model evaluates uncertainty by calculating the entropy of the predicted distribution for each unlabeled sample.
\\
$\circ$ \textbf{LC} selects samples with the lowest highest predicted probability for annotation. The computation is defined as
$  \hat{y} = \arg\max_y P(y|x)$, where $\hat{y}$ denotes the highest probability the model assigns to any category for the input $x$.
\\
{
$\circ$ \textbf{BALD} introduces Bayesian AL by Disagreement. It selects data points that maximize the mutual information between the predicted labels and the posterior distributions of the model.}
\\
{
$\circ$ \textbf{Core-set} presents a core-set-based AL strategy to construct a core subset that minimizes the maximum distance between any unlabeled sample and its nearest selected center.}
\\
$\circ$ \textbf{BERT-KM} applies k-means clustering on the L2-normalized BERT embedding outputs to select the cluster centers for sampling.
\\
$\circ$ \textbf{Patron} utilizes hard prompts, employing a prompt-based uncertainty propagation method to evaluate sample significance and a partition-then-rewrite (PTR) technique to improve diversity.
\\
$\circ$ \textbf{CAL} selects samples that are similar within the feature space of the model while exhibiting the most diverse predictive probabilities.
\\
$\circ$ \textbf{BADGE} measures the informativeness based on the parameter gradients of the final output layer and selects a batch of samples whose gradients display diverse directional orientations.
\\
$\circ$ \textbf{Random} randomly selects samples from the unlabeled pool.
%
\begin{figure*}[!bht]
    \centering
    \begin{subfigure}{0.25\textwidth}
        \includegraphics[width=\textwidth]{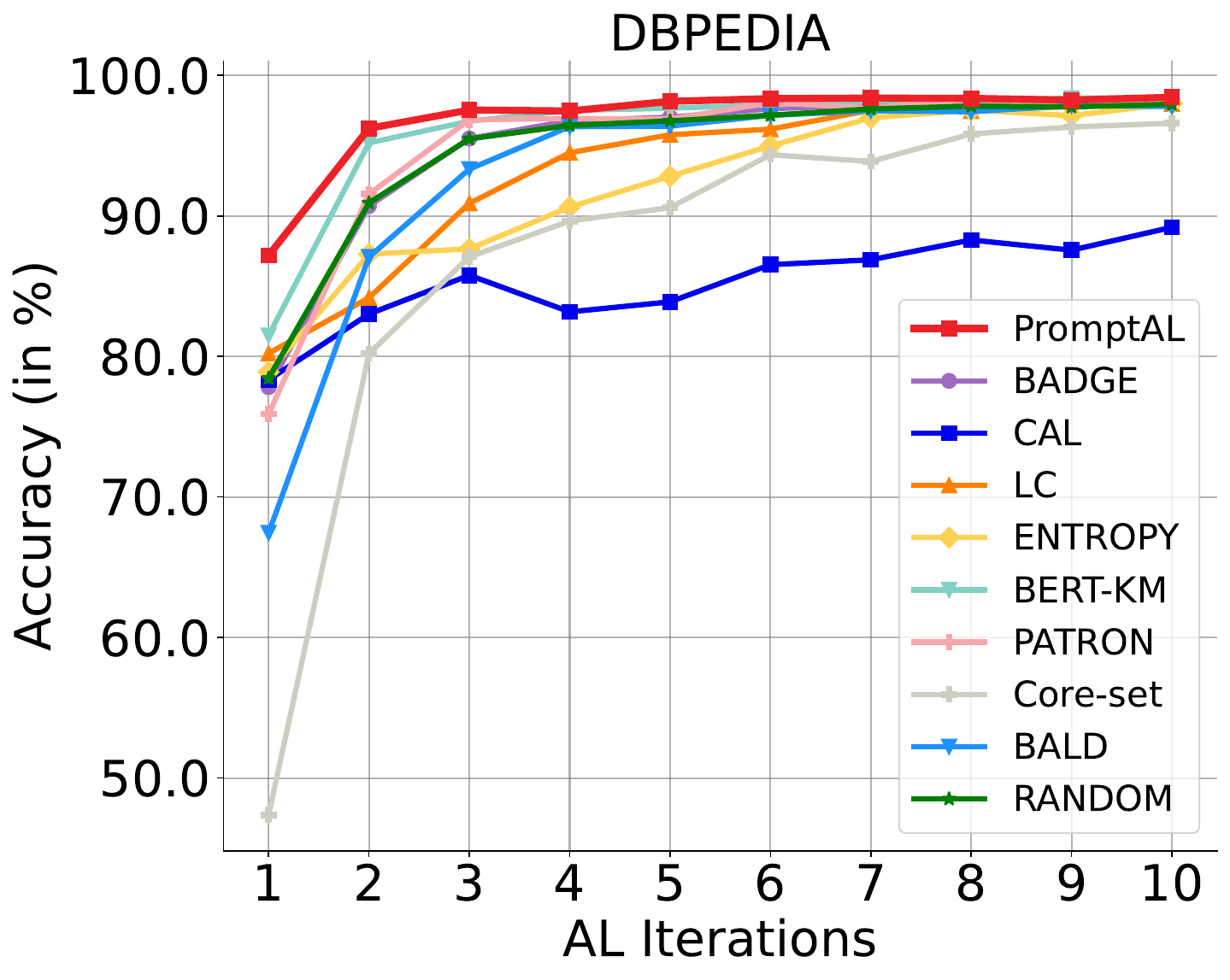}
    \end{subfigure}
    \hspace{0.03\textwidth} 
    \begin{subfigure}{0.25\textwidth}
        \includegraphics[width=\textwidth]{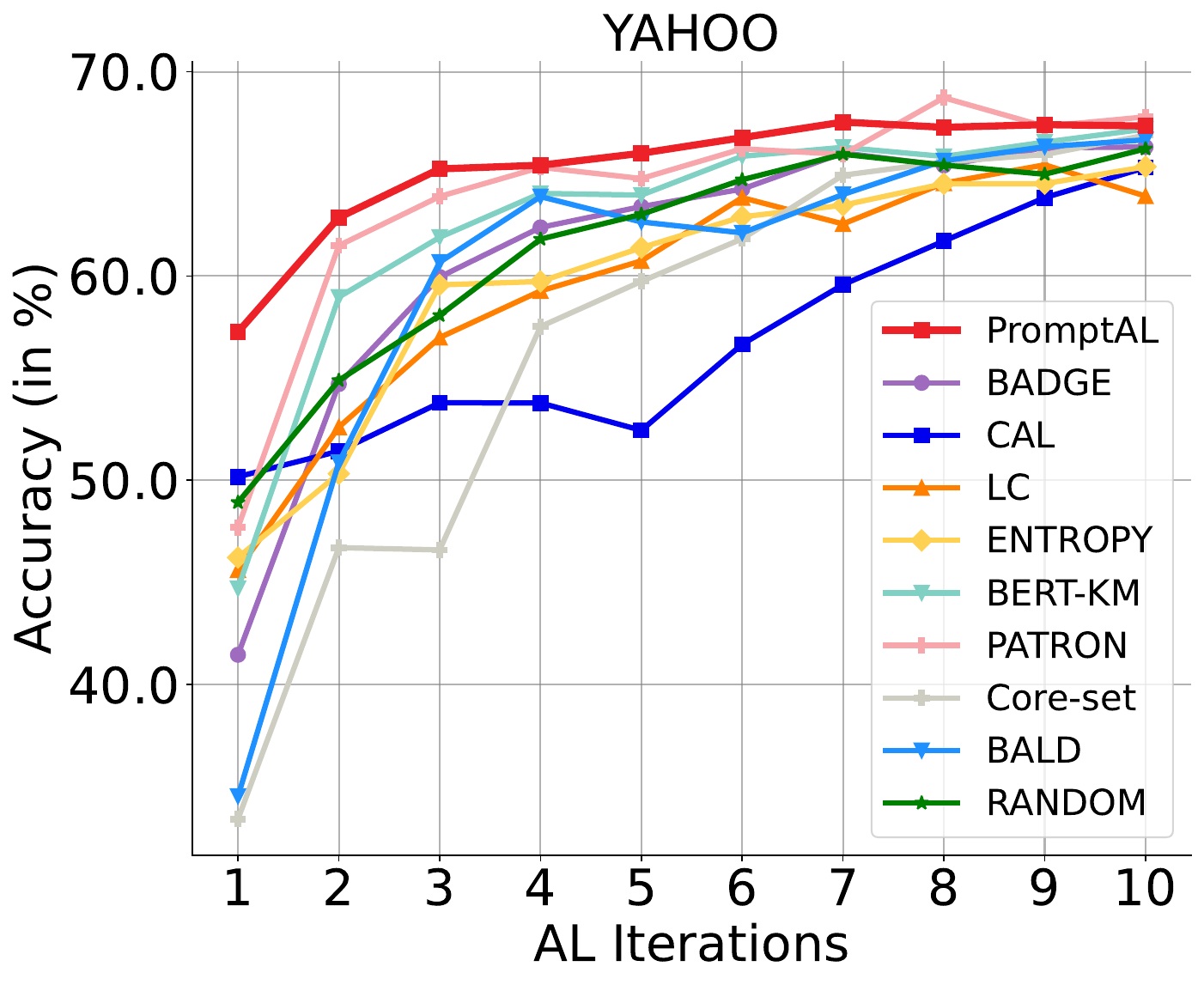}
    \end{subfigure}
    \hspace{0.03\textwidth} 
    \begin{subfigure}{0.25\textwidth}
        \includegraphics[width=\textwidth]{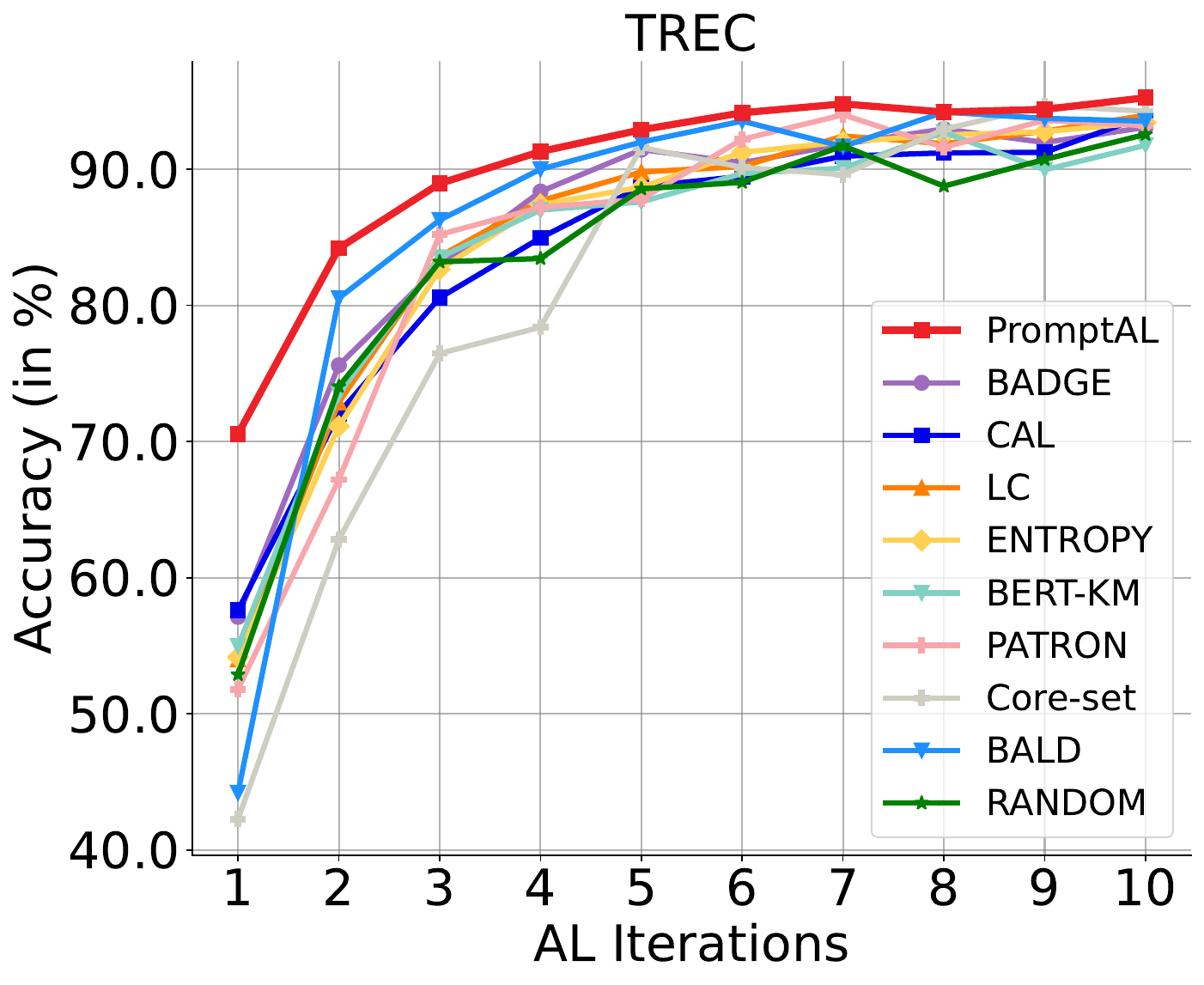}
    \end{subfigure}
    \\
    \vspace{0.00cm} 
    \begin{subfigure}{0.25\textwidth}
        \includegraphics[width=\textwidth]{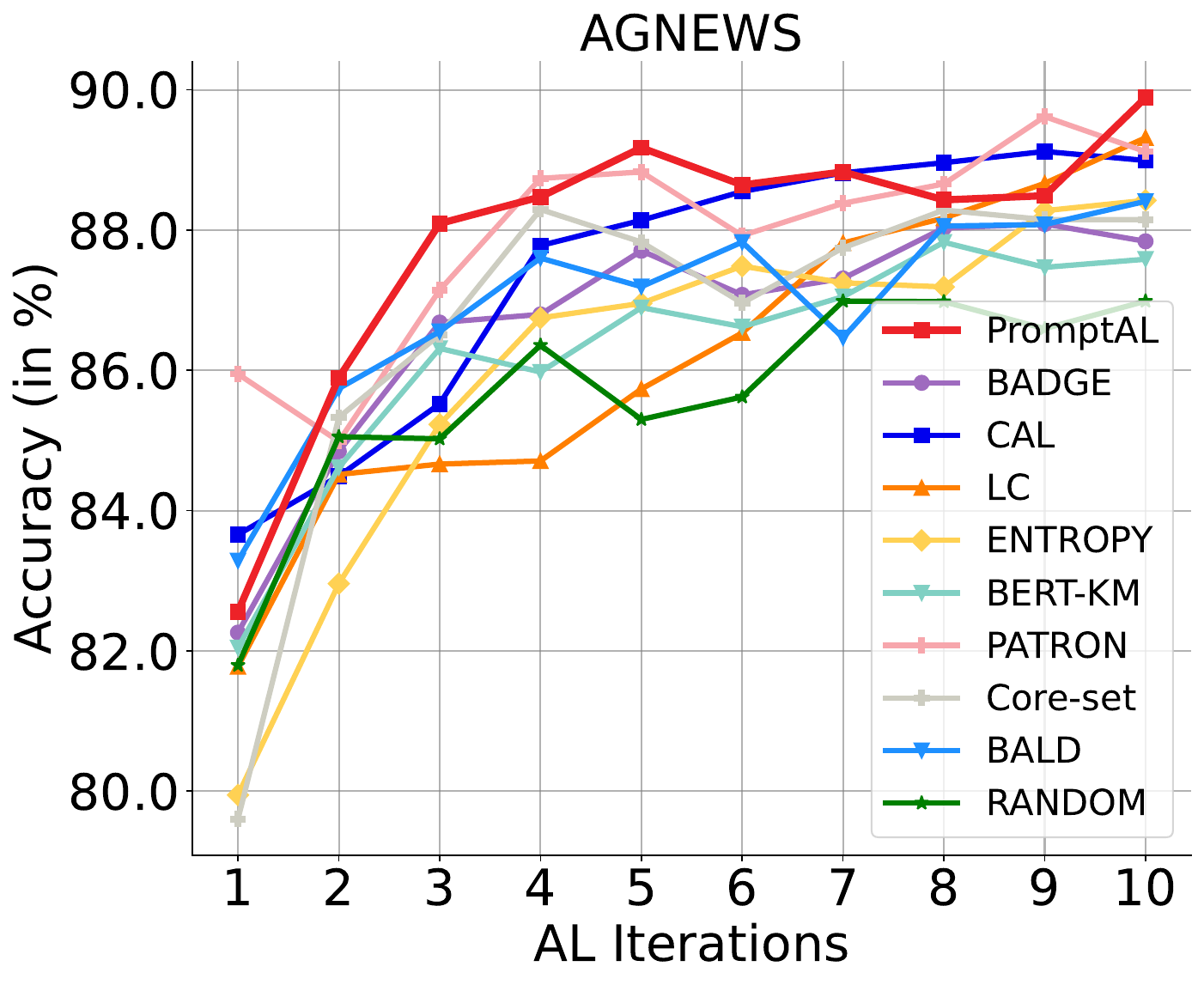}
    \end{subfigure}
    \hspace{0.03\textwidth} 
    \begin{subfigure}{0.25\textwidth}
        \includegraphics[width=\textwidth]{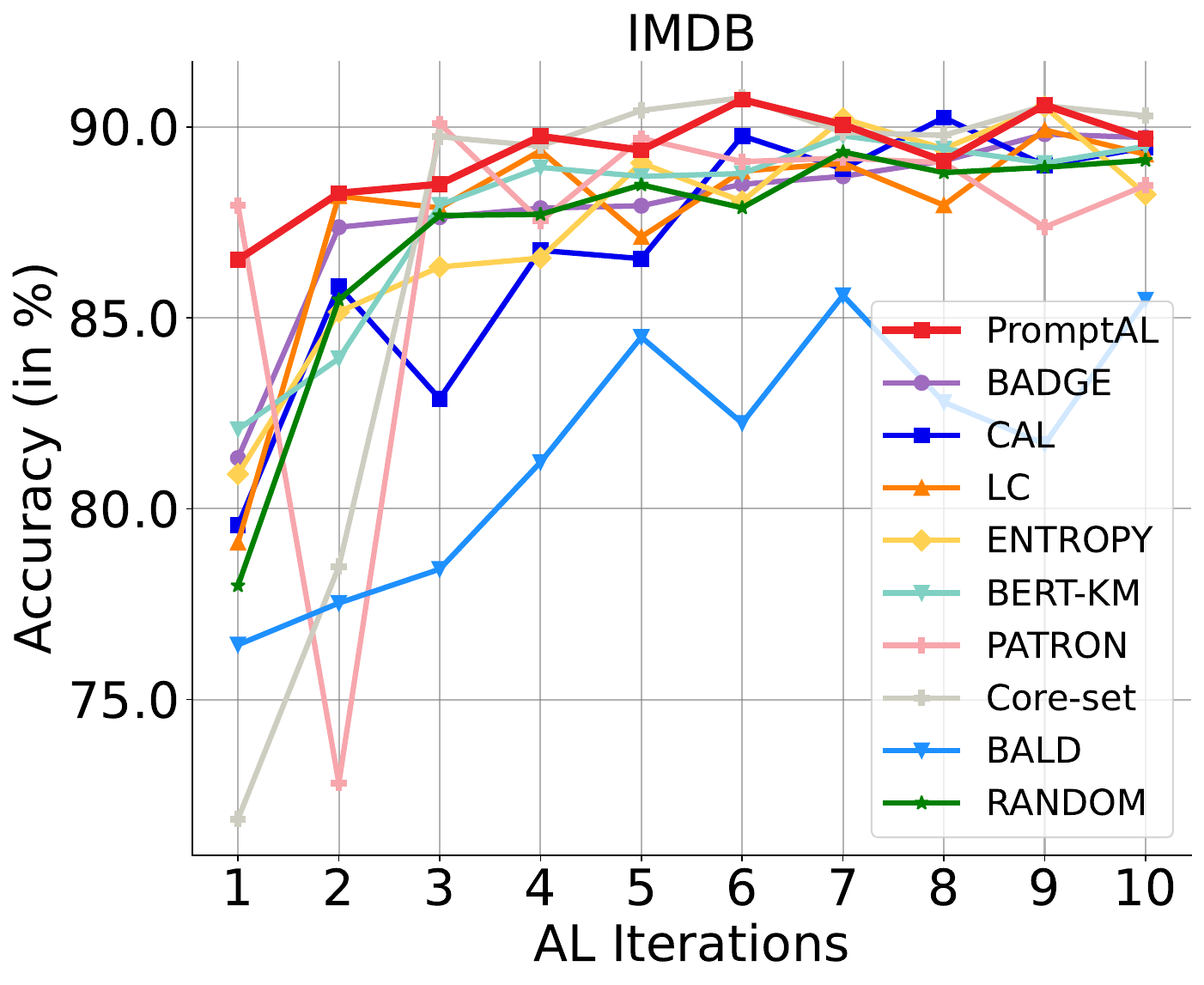}
    \end{subfigure}
    \hspace{0.03\textwidth} 
    \begin{subfigure}{0.25\textwidth}
        \includegraphics[width=\textwidth]{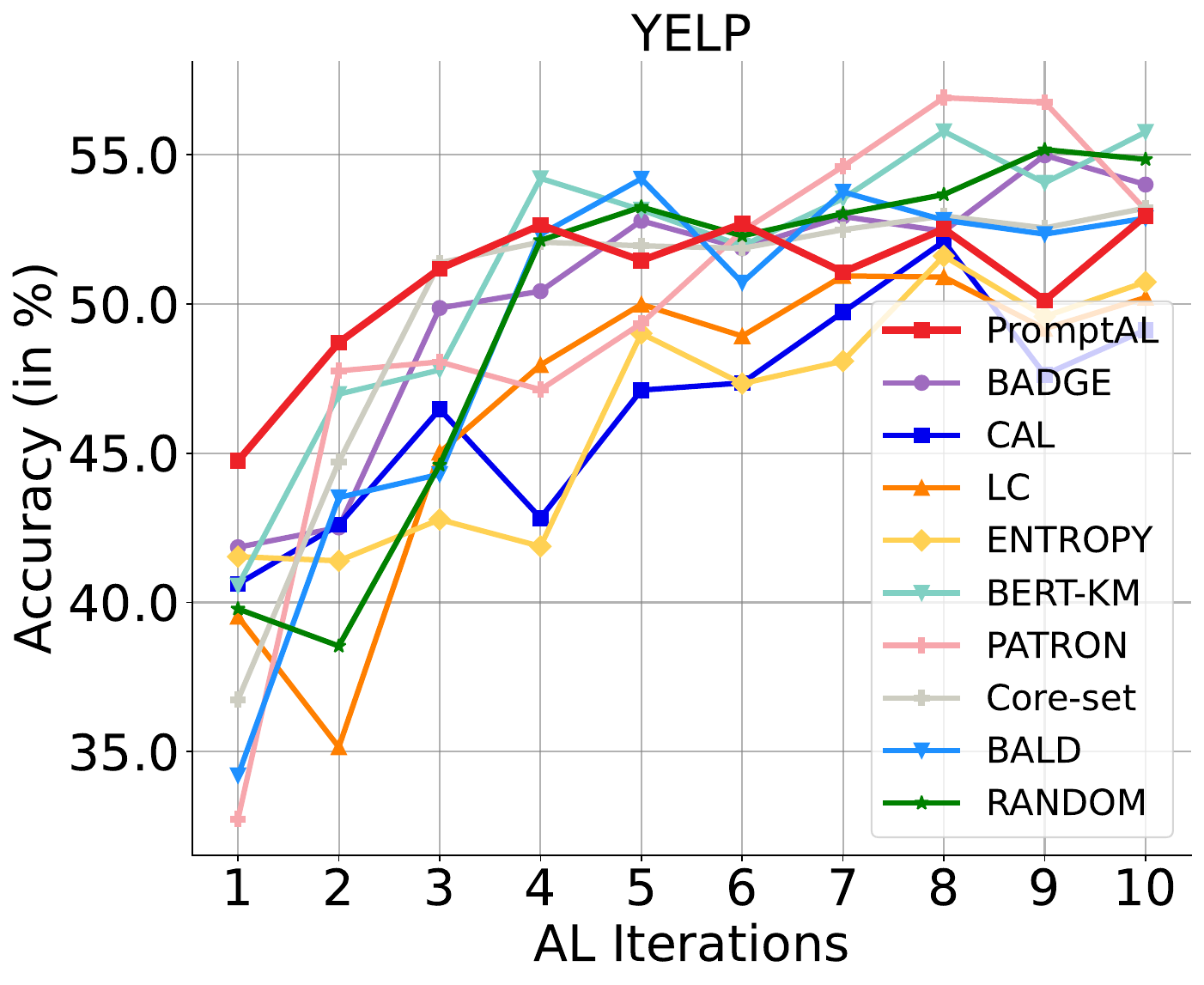}
    \end{subfigure}
    
    \caption{In-domain test accuracy during AL iterations for different query strategies.}
    \label{fig:acc}
\end{figure*}
\subsection{Main Results}
\label{Main Results}
Figure \ref{fig:acc} illustrates the in-domain test accuracy of eight AL methods across six datasets.
The study highlights the following key findings:

(1) PromptAL consistently outperforms all other methods across all six tasks.
Notably, on the DBpedia dataset, PromptAL surpasses competitors by an average margin of \textbf{3.56\%}.
This demonstrates the benefits of utilizing sample-aware dynamic soft prompts to fine-tune decision boundaries.
By refining these boundaries, selecting samples with higher distribution representativeness markedly enhances AL method performance.

(2) PromptAL achieves greater accuracy compared to the Patron method, which employs hard prompts.
For instance, on the DBpedia dataset, PromptAL realizes an average accuracy improvement of \textbf{2.01\%}.
Unlike fixed hard prompts, the soft prompts of PromptAL enable dynamic and efficient distribution adjustments, aligning more closely with the target distribution and thereby improving AL performance.

(3) The performance gains from PromptAL are more significant in datasets with a larger number of categories.
On the Yahoo, DBpedia, and AGNews datasets, PromptAL attains comparable performance in just \textbf{3} iterations, whereas the Entropy method requires \textbf{10} iterations to reach similar effectiveness.
This observation demonstrates that PromptAL effectively reduces the number of iterations and enhances efficiency by identifying samples that exhibit more representative characteristics of their respective categories.

(4) In few-shot settings, uncertainty-based techniques (e.g., Entropy and LC) and diversity-based approaches (e.g., BERT-KM) frequently yield subpar results. Uncertainty-based methods omit diversity information, leading to the selection of samples that poorly represent the overall distribution. Conversely, diversity-based approaches, lacking uncertainty estimates, may bias decision boundaries and select less informative, easier samples. In contrast, PromptAL combines diversity and uncertainty, effectively choosing samples that are both distribution-representative and highly informative.
%
%

Overall, PromptAL demonstrates superior performance by utilizing sample-aware dynamic soft prompts, which adjust the distribution to better align with the target, making it a more effective solution for few-shot AL.
\subsection{Statistical Significance Analysis}
{ To evaluate the statistical significance of the performance of the PromptAL, paired t-tests were conducted across six datasets. Specifically, for each method, experiments were run with five random seeds. Each seed underwent ten rounds of AL on each dataset, and the average accuracy across the ten runs was calculated as the final metric.
Paired t-tests were then performed between PromptAL and each baseline on each dataset using these final results from five different random seeds.
We present the average accuracy of PromptAL and baselines across ten AL rounds in Table \ref{acc}, along with the corresponding paired t-test results in Table \ref{p}.
}
\begin{table}[h]
\centering
\caption{{Average accuracy (\%) of PromptAL and baselines over ten rounds. The best results are in bold.}}
\resizebox{0.8\textwidth}{!}{
\begin{tabular}{lcccccc}
\toprule
Method & AGNews & DBPedia & IMDB & TREC & Yahoo & Yelp \\
\midrule
BERT-KM     & 85.17 & 93.72 & 86.32 & 77.31 & 58.71 & 48.54 \\
BADGE       & 85.66 & 91.57 & 86.43 & 79.08 & 56.38 & 47.49 \\
CAL         & 85.92 & 82.84 & 84.32 & 76.78 & 52.32 & 43.93 \\
LC         & 84.28 & 89.11 & 86.34 & 77.59 & 55.04 & 43.53 \\
Entropy     & 84.37 & 87.46 & 85.60 & 76.80 & 55.45 & 43.32 \\
Patron      & \textbf{87.13} & 91.63 & 85.62 & 75.84 & 60.63 & 45.01 \\
Core-set    & 85.51    & 78.98   & 84.00    & 70.31 & 48.79 & 48.06    \\  
BALD        & 86.07 & 88.12 & 79.62 & 78.60 & 54.52 & 45.71 \\

Random      & 84.71 & 91.61 & 85.47 & 76.42 & 57.34 & 45.66 \\
PromptAL    & 86.84 & \textbf{95.32} & \textbf{88.49} & \textbf{85.46} & \textbf{63.36} & \textbf{49.75} \\
\bottomrule
\end{tabular}
}
\label{acc}
\end{table}
%
%
%
\begin{table}[htb]
\centering
\caption{{Paired \textit{t}-test results ($p$-values) between PromptAL and baselines on six datasets. \cmark\ indicates $p < 0.05$ (statistically significant), while \xmark\ indicates $p \geq 0.05$ (not statistically significant).}}
\resizebox{0.8\textwidth}{!}{
\begin{tabular}{lcccccc}
\toprule
Method & AGNews & DBPedia & IMDB & TREC & Yahoo & Yelp \\
\midrule
BERT-KM     & 0.024\cmark & 0.178\xmark & 0.101\xmark & 0.002\cmark & 0.009\cmark & 0.194\xmark \\
BADGE       & 0.159\xmark & 0.002\cmark & 0.042\cmark & 0.004\cmark & 0.000\cmark & 0.049\cmark \\
CAL         & 0.270\xmark & 0.000\cmark & 0.050\cmark & 0.005\cmark & 0.000\cmark & 0.021\cmark \\
LC          & 0.042\cmark & 0.005\cmark & 0.034\cmark & 0.001\cmark & 0.002\cmark & 0.020\cmark \\
Entropy     & 0.013\cmark & 0.002\cmark & 0.029\cmark & 0.001\cmark & 0.002\cmark & 0.004\cmark \\
Patron      & 0.544\xmark & 0.008\cmark & 0.033\cmark & 0.050\cmark & 0.048\cmark & 0.016\cmark \\
Core-set    & 0.005\cmark & 0.000\cmark & 0.044\cmark & 0.007\cmark & 0.006\cmark & 0.039\cmark \\
BALD        & 0.028\cmark & 0.002\cmark & 0.000\cmark & 0.020\cmark & 0.004\cmark & 0.002\cmark \\
Random      & 0.016\cmark & 0.004\cmark & 0.007\cmark & 0.024\cmark & 0.000\cmark & 0.024\cmark \\
\bottomrule
\end{tabular}
}\label{p}
\end{table}

{ As listed in Table \ref{acc}, the PromptAL achieves the highest average accuracy across all datasets except AGNews, where it ranks second.
Furthermore, as illustrated in Table \ref{p}, PromptAL exhibits statistical significance over eight of the nine baselines—excluding only BERT-KM—across the DBPedia, TREC, Yahoo, and Yelp datasets.
Additionally, Table~\ref{acc} shows that PromptAL attains the highest accuracy across these datasets, suggesting meaningful performance enhancements.
This underscores meaningful performance enhancements, as PromptAL attains the top accuracy on these datasets.
However, on the AGNews and IMDB datasets, PromptAL does not reach statistical significance compared to four baselines: BADGE, BERT-KM, Patron, and CAL. Despite this, its accuracy remains comparable to these methods, likely due to the limited number of classes in these datasets.
Since all methods, including PromptAL, can easily classify these datasets, PromptAL does not exhibit a distinct advantage.
Conversely, PromptAL demonstrates more substantial benefits on datasets with a greater number of classes, such as TREC and DBPedia, which have 6 and 14 classes, respectively.
Overall, PromptAL shows significant improvements across most tasks, affirming its effectiveness in few-shot scenarios.}

\subsection{Out-of-Distribution Analysis}
\label{ood_Analysis}
We present the OOD performance of {ten} query strategies after ten iterations of AL in Table \ref{ood}.
{Overall, PromptAL outperforms the latest method, Patron, by \textbf{3.16\%} across all three OOD datasets. Notably, PromptAL achieves the highest accuracy on both the SST-2 and IMDB-Contrast datasets and secures the second position on the IMDB-Counter dataset.}
This indicates that PromptAL demonstrates superior generalization capabilities.
We found that hybrid and diversity-based methods generally surpass uncertainty-based methods, suggesting that diverse samples are more advantageous for generalization in low-resource scenarios.
PromptAL further enhances generalization by incorporating diversity and utilizing sample-aware information from OOD data to adjust predictive distributions, leading to superior performance.
\begin{table}[t]
    \centering
    \caption{OOD accuracy (in \%) for IMDB across AL methods.}
\resizebox{0.8\textwidth}{!}{
    \begin{tabular}{cccc}
        \toprule
        Train (ID) & IMDB & IMDB & IMDB  \\
        Test (OOD) & SST-2 & IMDB-Contrast & IMDB-Counter\\
        \midrule
        Random & 83.7 ± 3.3 & 88.3 ± 0.9 & 91.2 ± 3.3    \\
       BERT-KM & 86.4 ± 1.6 &	87.8 ± 3.3&	89.4 ± 2.6       \\
        {Core-set}& {83.5 ± 0.5} &	{88.9 ± 1.0}&	{93.0 ± 0.6}       \\
       { BALD }& {85.5 ± 1.5} &	{85.0 ± 1.9}&	{87.9 ± 0.8 }      \\
       Entropy&	81.4 ± 0.8&	89.0 ± 1.1&	91.7 ± 1.7      \\
        LC &	82.8 ± 3.0&	86.2 ± 3.1&	92.5 ± 1.8      \\
         CAL	&85.9 ± 2.5	&86.7 ± 3.9	&86.4 ± 8.6     \\
         BADGE&	86.2 ± 4.1&	87.0 ± 1.8&	92.4 ± 0.6 \\
         Patron&	86.4 ± 1.0&	84.2 ± 1.8&	89.8 ± 1.9 \\
         \textbf{PromptAL}& 	\textbf{87.3 ± 0.8}&	\textbf{89.8 ± 1.9}&	\textbf{92.8 ± 1.3} \\
        \bottomrule
    \end{tabular}
    }
    \label{ood}
\end{table}
\subsection{Ablation Study}
\label{Ablation Study}
We conduct extensive ablation studies and present the results for the TREC and AGNews datasets, focusing on three key aspects: (1) the ablation of PromptAL modules, (2) the ablation of the fusion mechanism, and (3) the ablation of the feature space in global diversity.
\subsubsection{Ablation of PromptAL Modules}
\begin{figure*}[!bht]
    \centering
    \begin{subfigure}{0.3\textwidth}
        \includegraphics[width=\textwidth]{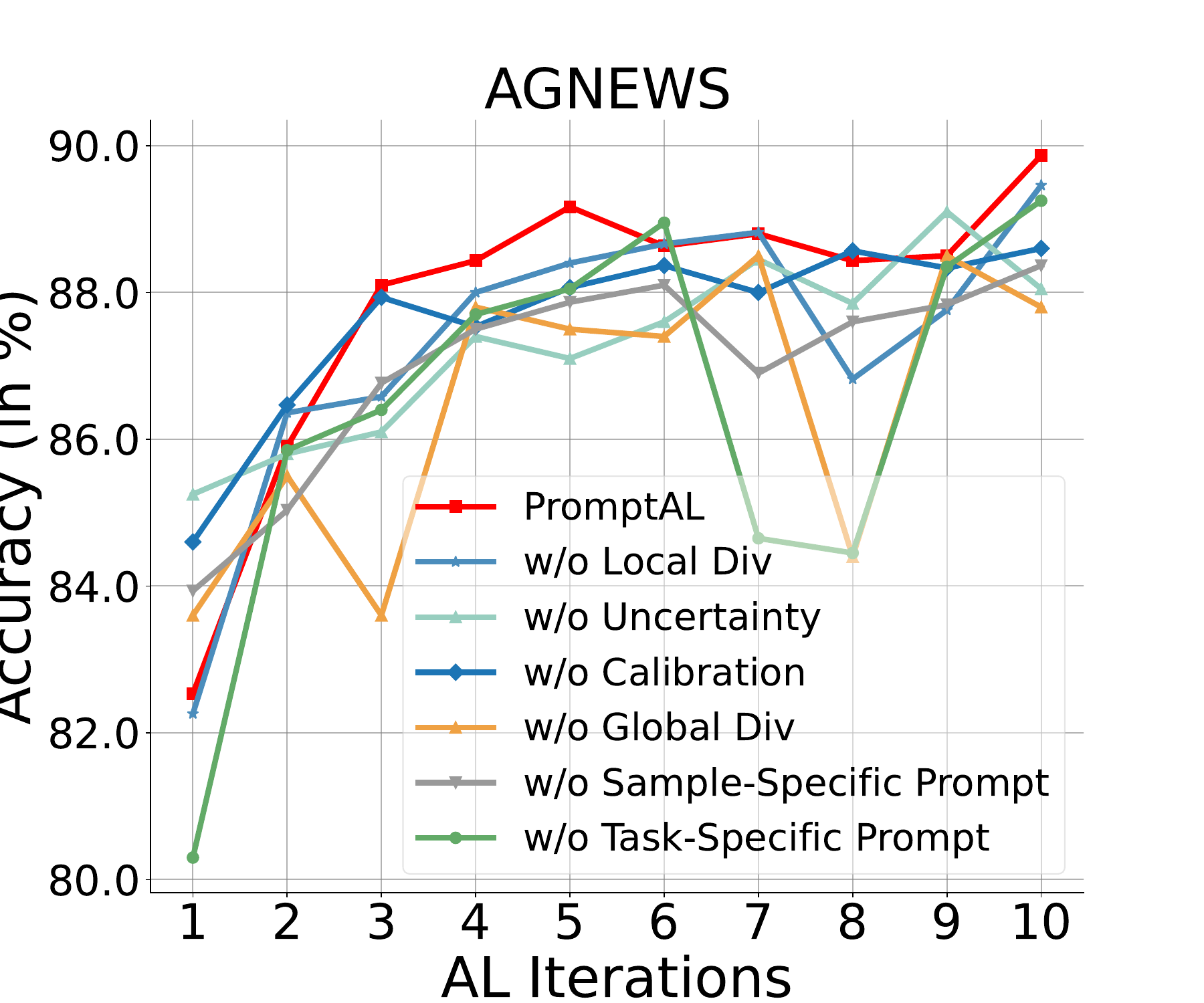}
    \end{subfigure}
    \begin{subfigure}{0.3\textwidth}
        \includegraphics[width=\textwidth]{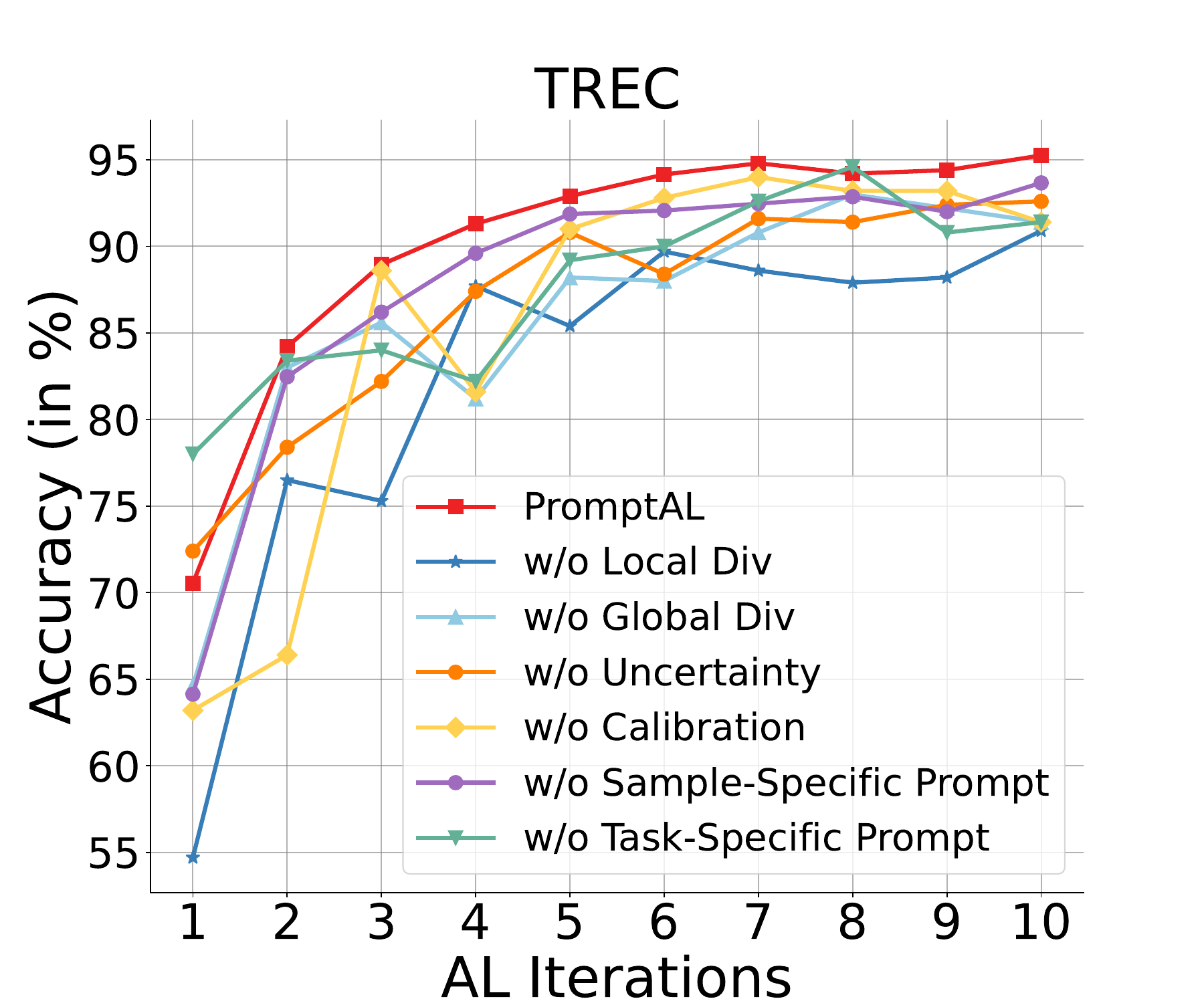}
    \end{subfigure}

    \caption{Results of the PromptAL module
ablation analysis.}
    \label{ablation}
\end{figure*}
We comprehensively analyzed the contribution of each component to PromptAL, with results shown in Fig.\ref{ablation}, and drew the following conclusions.

(1) Without sample-specific prompts, the prompt reverts to a fixed soft prompt after training, and the model's overall performance declines.
This demonstrates that sample-specific prompts, which exert varying effects depending on the samples, are essential for steering the model's improvement toward the target distribution.

(2) Without considering task-specific prompts shared across all samples, the improvement in model performance is significantly diminished.
This demonstrates that these prompts enhance the task comprehension of the model and promote its generalizability.

(3) Both global and local diversity are pivotal in influencing performance.
The absence of either leads to a substantial decline and instability in the performance of the model, confirming that diverse samples are crucial for model improvement in few-shot scenarios.

(4) Without accounting for uncertainty, the model's performance diminishes significantly.
This indicates that PromptAL improves the quality of selected samples by appropriately integrating uncertainty through the joint scoring mechanism.

(5) Experiments without calibration reveal that uncalibrated uncertainty scores degrade AL performance.
Therefore, PromptAL must utilize a \textit{contextualized prior} to rectify the inherent prediction bias of the model.

In summary, each module of PromptAL is indispensable; the removal of any module results in varying degrees of AL performance degradation.

\subsubsection{Ablation of Fusion Mechanism}
\begin{figure*}[!bht]
\centering
    \begin{subfigure}{0.3\textwidth}
        \includegraphics[width=\textwidth]{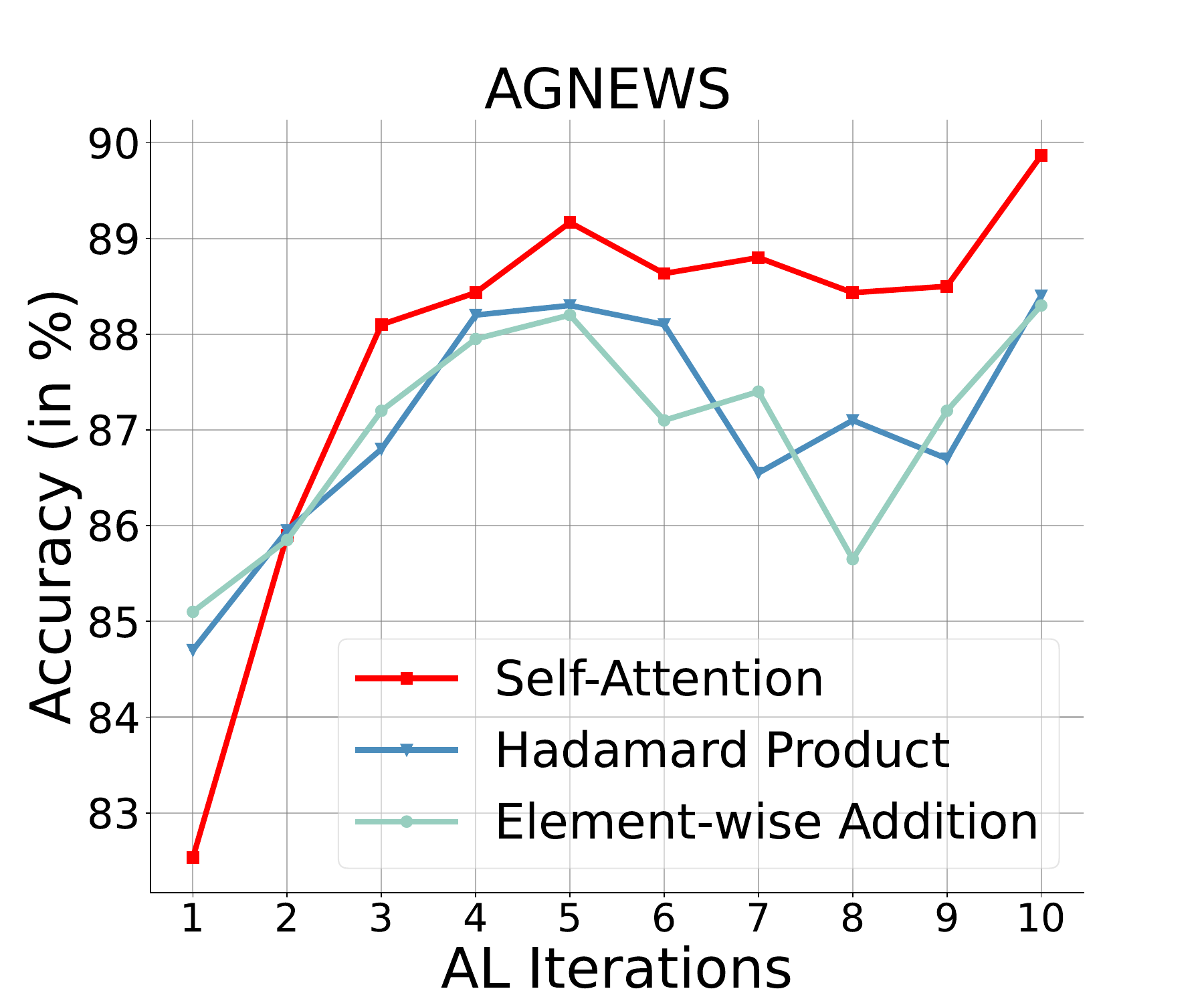}
    \end{subfigure}
    \begin{subfigure}{0.3\textwidth}
        \includegraphics[width=\textwidth]{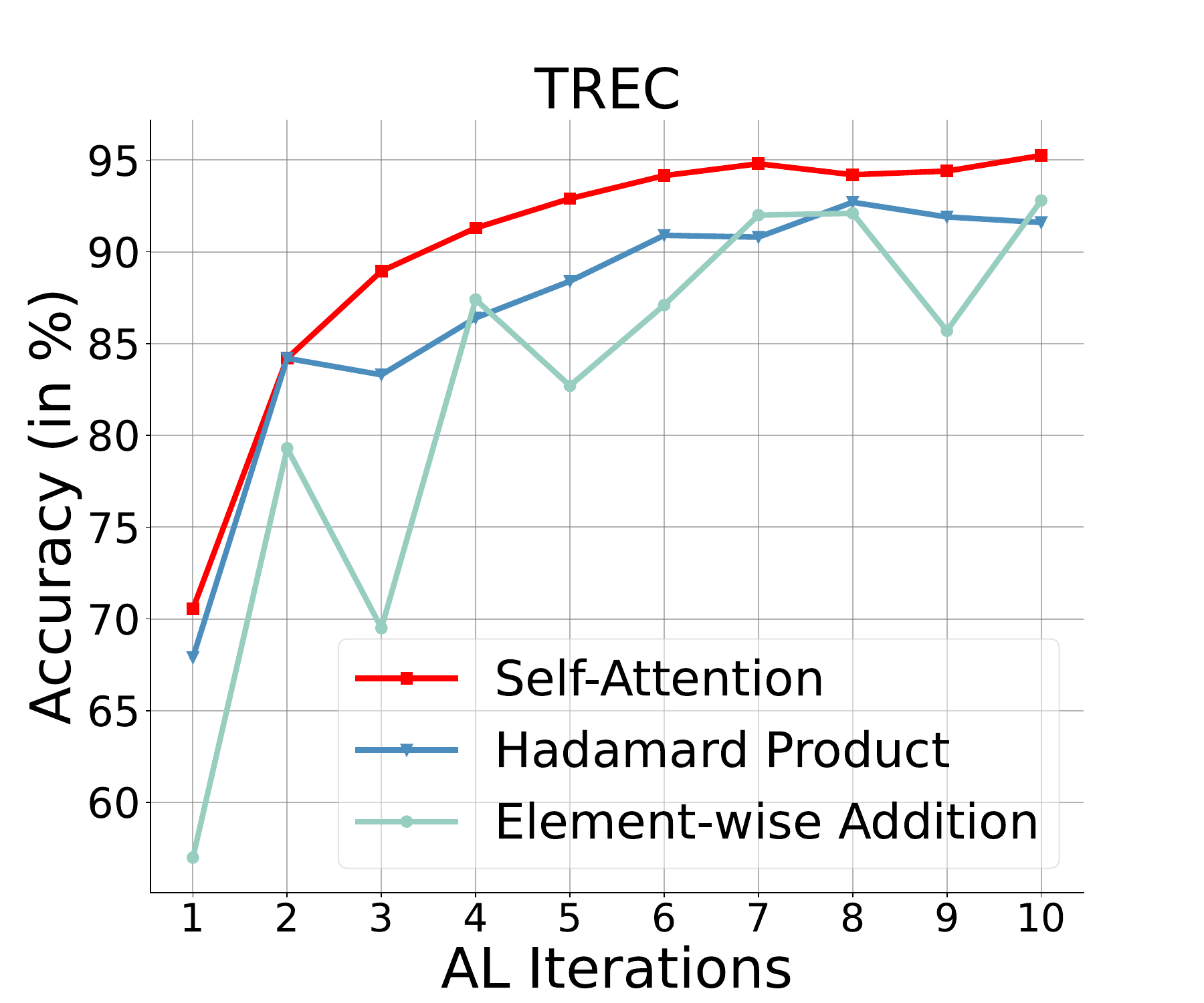}
    \end{subfigure}
    \caption{Results of different fusion mechanisms.}
       \label{combine}
\end{figure*}
We assessed different fusion mechanisms for integrating task and sample prompts.
In PromptAL, the self-attention mechanism is substituted with two alternatives: the \textbf{Hadamard Product} and \textbf{Element-wise Addition}.
In both approaches, a sigmoid function is applied to the sample-specific prompt to derive a coefficient, which is then merged with the task-specific prompt.
As illustrated in Fig.\ref{combine}, the self-attention mechanism outperforms both the Hadamard product and element-wise addition.
We believe this is because the self-attention mechanism can dynamically filter and extract useful sample-aware information for positive fitting of the target distribution.
Conversely, the other two methods lack a filtering mechanism, thereby introducing noisy information that adversely impacts the task-specific prompt and degrades performance.
\subsubsection{Ablation of Feature Space }
\begin{figure*}[!bht]
\centering
    \begin{subfigure}{0.3\textwidth}
        \includegraphics[width=\textwidth]{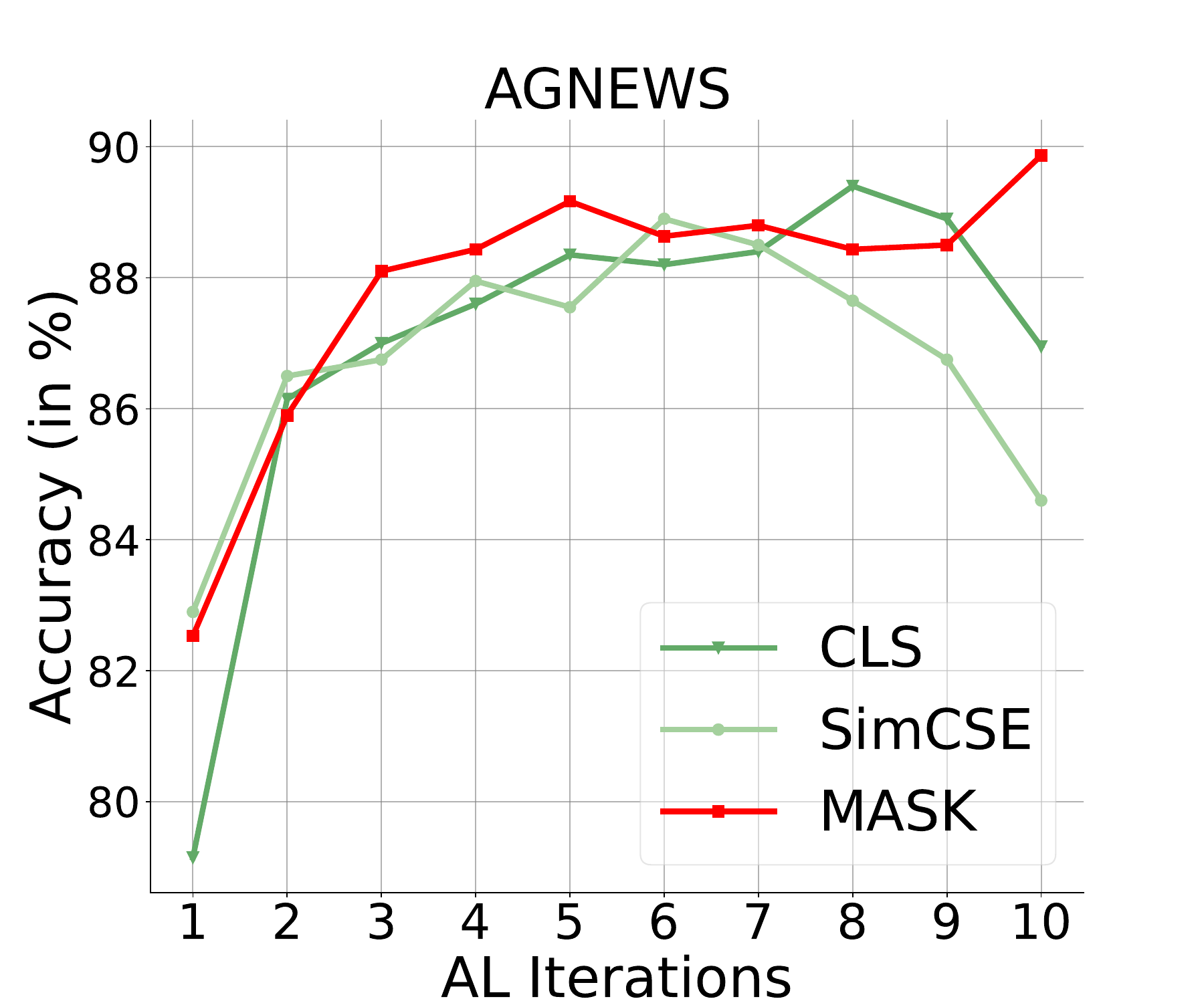}
    \end{subfigure}
    \begin{subfigure}{0.3\textwidth}
        \includegraphics[width=\textwidth]{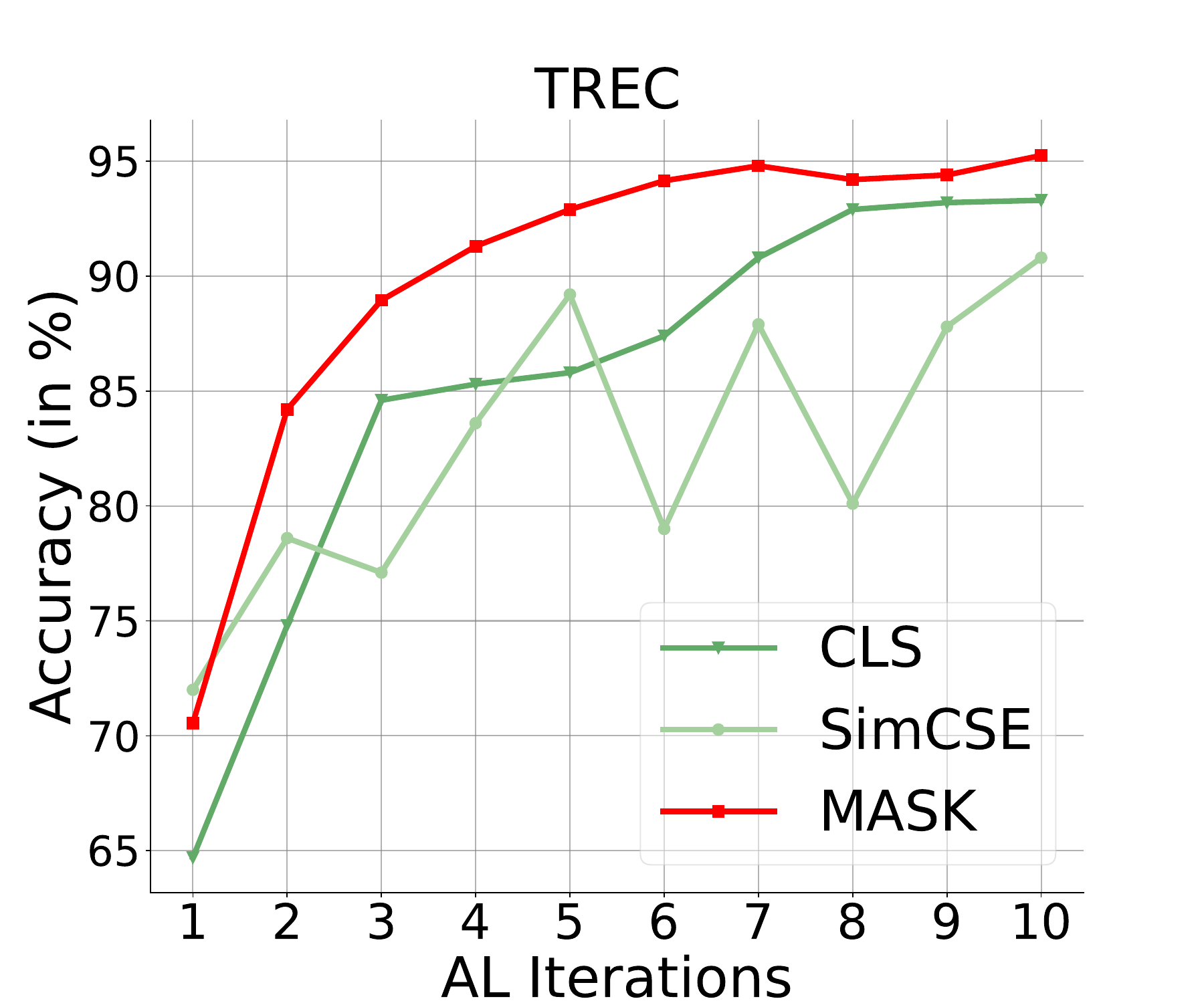}
    \end{subfigure}
    \caption{Results of different feature spaces for global diversity.}
      \label{cluster}
\end{figure*}
We conducted a comparative experiment on feature space representations by substituting the knowledge feature with two alternative representations:
 (1) the semantic feature at the \texttt{[CLS]} position of $\mathcal{M}$ and (2) the semantic feature at the \texttt{[CLS]} position derived from the SimCSE model \cite{simcse}.
As illustrated in Fig.\ref{cluster}, the knowledge feature significantly outperforms both the semantic feature at the \texttt{[CLS]} position and the one obtained from SimCSE.
This superior performance is attributed to the knowledge feature's ability to capture task-related information from PLMs through dynamic soft prompts, thereby providing more comprehensive and diverse data than traditional semantic features and enhancing feature space modeling in PromptAL.

\subsection{Hyperparameter Analysis}
\label{Hyperparameter Analysis}
\begin{figure*}[htp]
    \begin{minipage}{0.23\textwidth} 
        \begin{subfigure}[b]{0.99\textwidth} 
    \includegraphics[width=\textwidth]{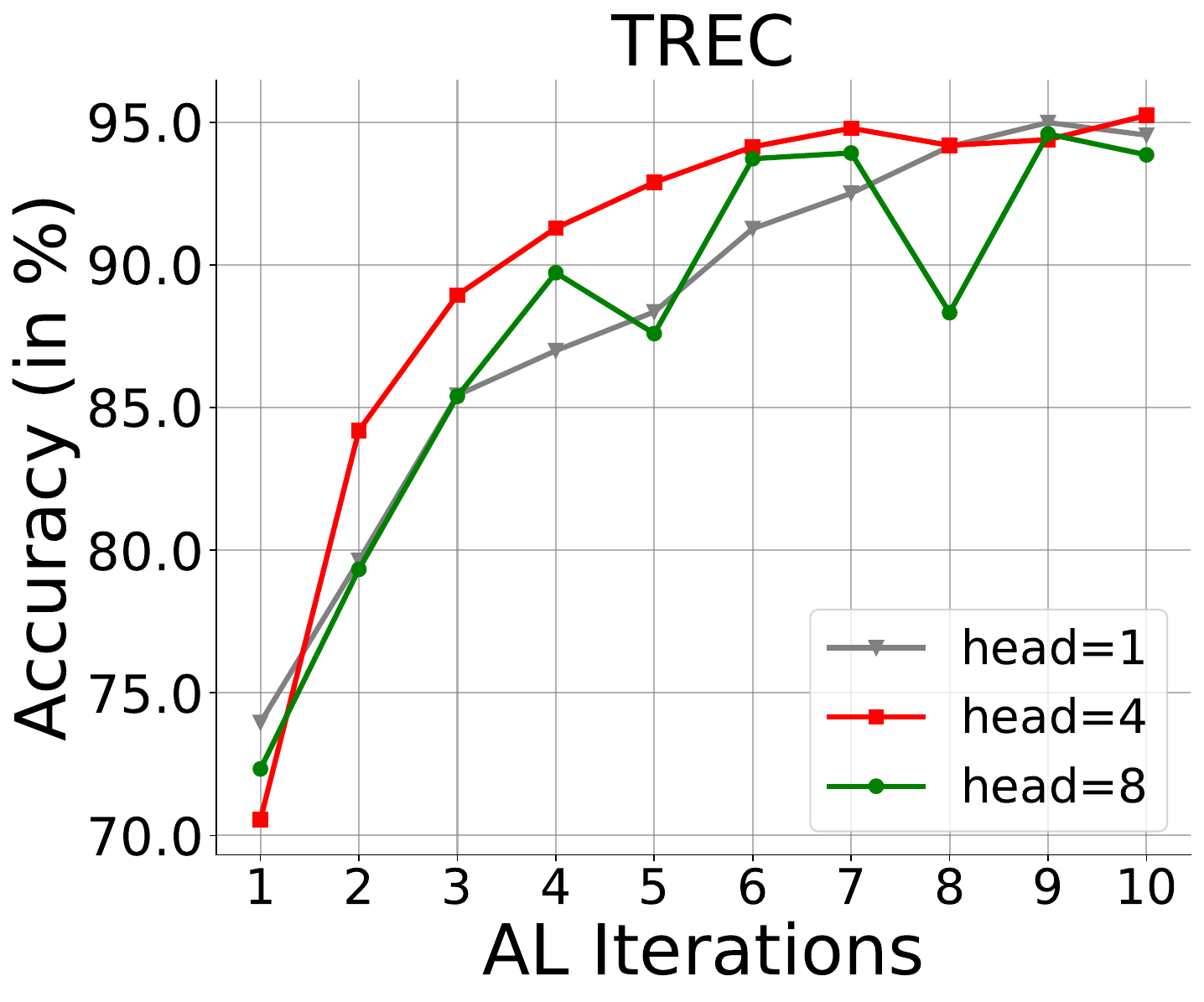}
        \end{subfigure}
        \caption{Impact of attention head numbers.}
        \label{head}
    \end{minipage}
    \begin{minipage}{0.49\textwidth} 
        \begin{subfigure}[b]{0.49\textwidth} 
            \includegraphics[width=\textwidth]{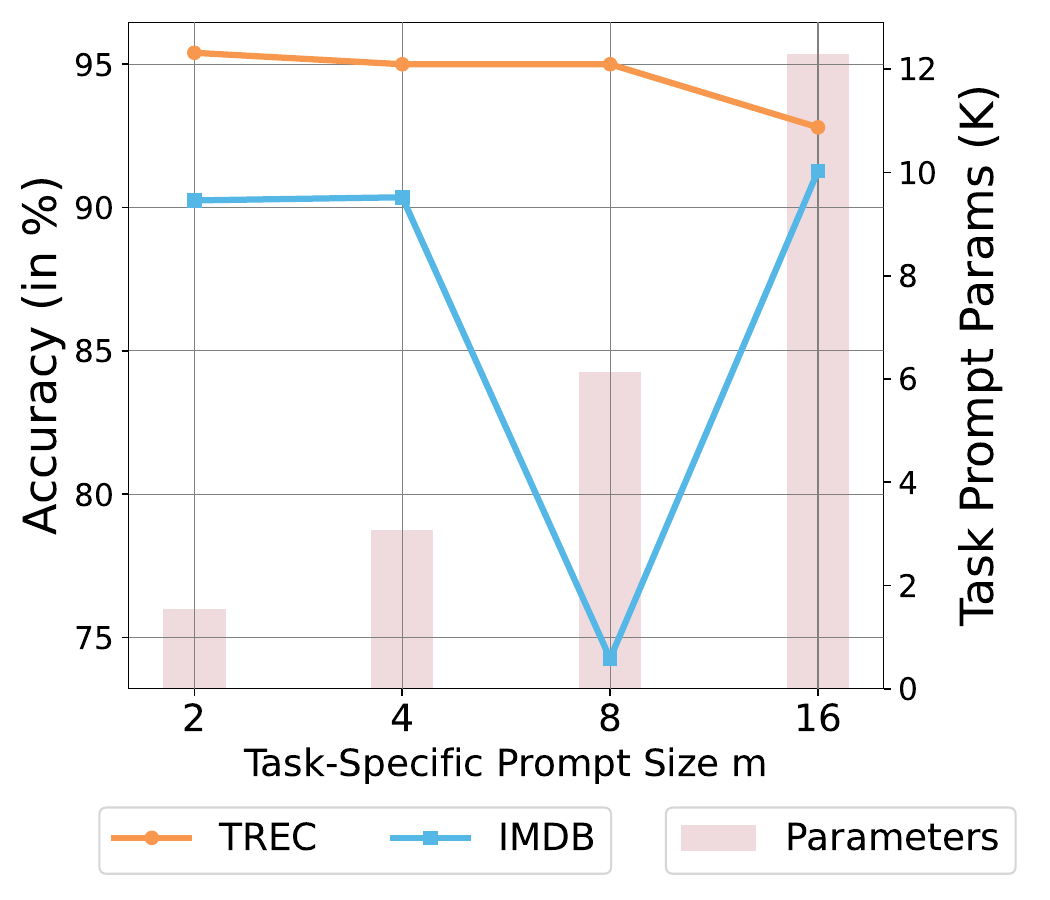}
          
            \label{fig:prompt_size_param_count}
        \end{subfigure}
        \begin{subfigure}[b]{0.49\textwidth} 
            \includegraphics[width=\textwidth]{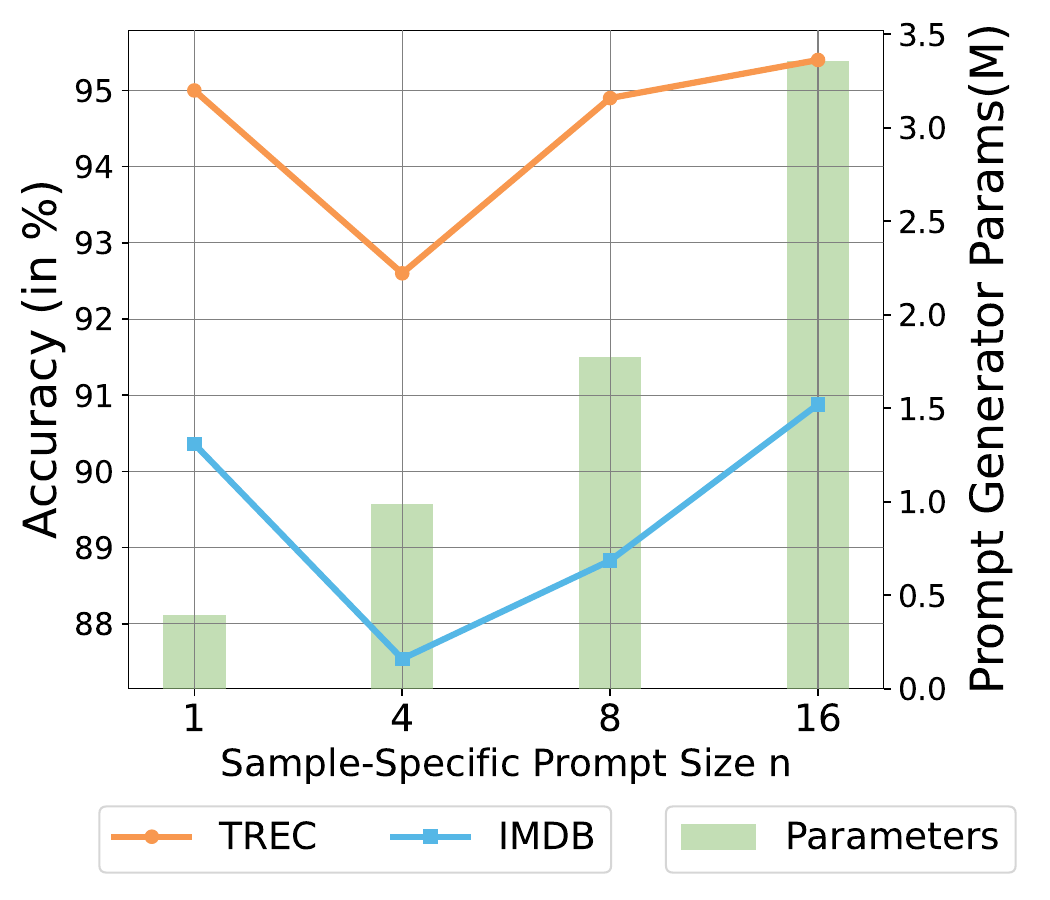}
        \end{subfigure}
        \caption{Results of different prompt sizes.}
        \label{size}
    \end{minipage}
    \hfill 
    \begin{minipage}{0.24\textwidth} 
        \begin{subfigure}[b]{0.99\textwidth} 
            \includegraphics[width=\textwidth]{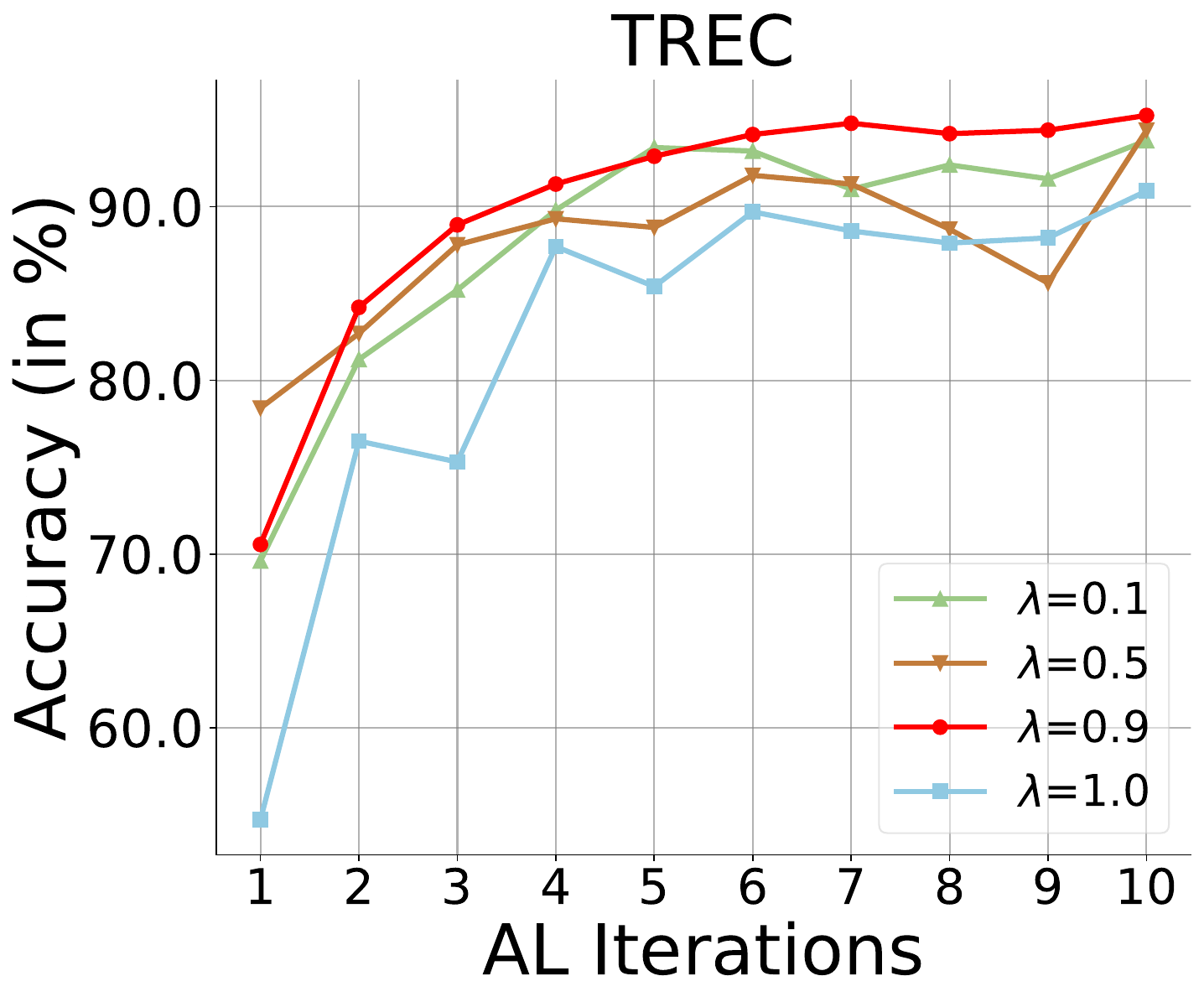}
        \end{subfigure}
        \caption{Impact of different $\lambda$.}
        \label{lambda}
    \end{minipage}
\end{figure*}
Additionally, we examined the impact of hyperparameters on AL performance, including the number of heads in the self-attention mechanism, the size $m$ of the task-specific prompt, the size $n$ of the sample-specific prompt, and the joint weight $\lambda$.
The experiments were conducted using 1, 4, and 8 multi-heads, with results presented in Fig.\ref{head}. 
The model achieves optimal performance when configured with 4 attention heads.
If the number of heads becomes excessively large (e.g., eight), accuracy exhibits fluctuations and diminished stability, primarily due to the addition of redundant parameters. Consequently, four heads emerge as the most effective configuration.

We conducted grid searches for the task-specific prompt size \(m\) and the sample-specific prompt size \(n\) within the ranges \{2, 4, 8, 16\} and \{1, 4, 8, 16\}, respectively.
We identified that \(m=4\) and \(n=1\) constitute a parameter-efficient and high-performing combination.
By fixing either \(m=4\) or \(n=1\), we present the results across ten AL rounds while varying the size of the other prompt in Fig.\ref{size}.
As \(n\) increases, the number of parameters in the prompt generator \(f\) grows. Similarly, increasing \(m\) leads to a rise in the parameters of the task prompt vector. The excessive values for \(m\) or \(n\) may result in over-parameterization, leading to a suboptimal model fit.

We conducted a search for \(\lambda\), which controls the weight distribution of each component in joint scores, within the range \([0.1, 0.5, 0.9, 1.0]\), as illustrated in Figure \ref{lambda}.
As \(\lambda\) increases, the influence of the uncertainty score progressively exceeds that of the local diversity score, leading to an overall enhancement in model performance.
However, when \(\lambda = 1\) (completely disregarding local diversity), performance diminishes, highlighting the essential role of local diversity.
Therefore, \(\lambda = 0.9\) emerges as the optimal choice.

\subsection{Distribution Alignment}
\label{Distribution Alignment}
\begin{figure*}[!bht]
    \centering
    \includegraphics[width=0.5\textwidth]{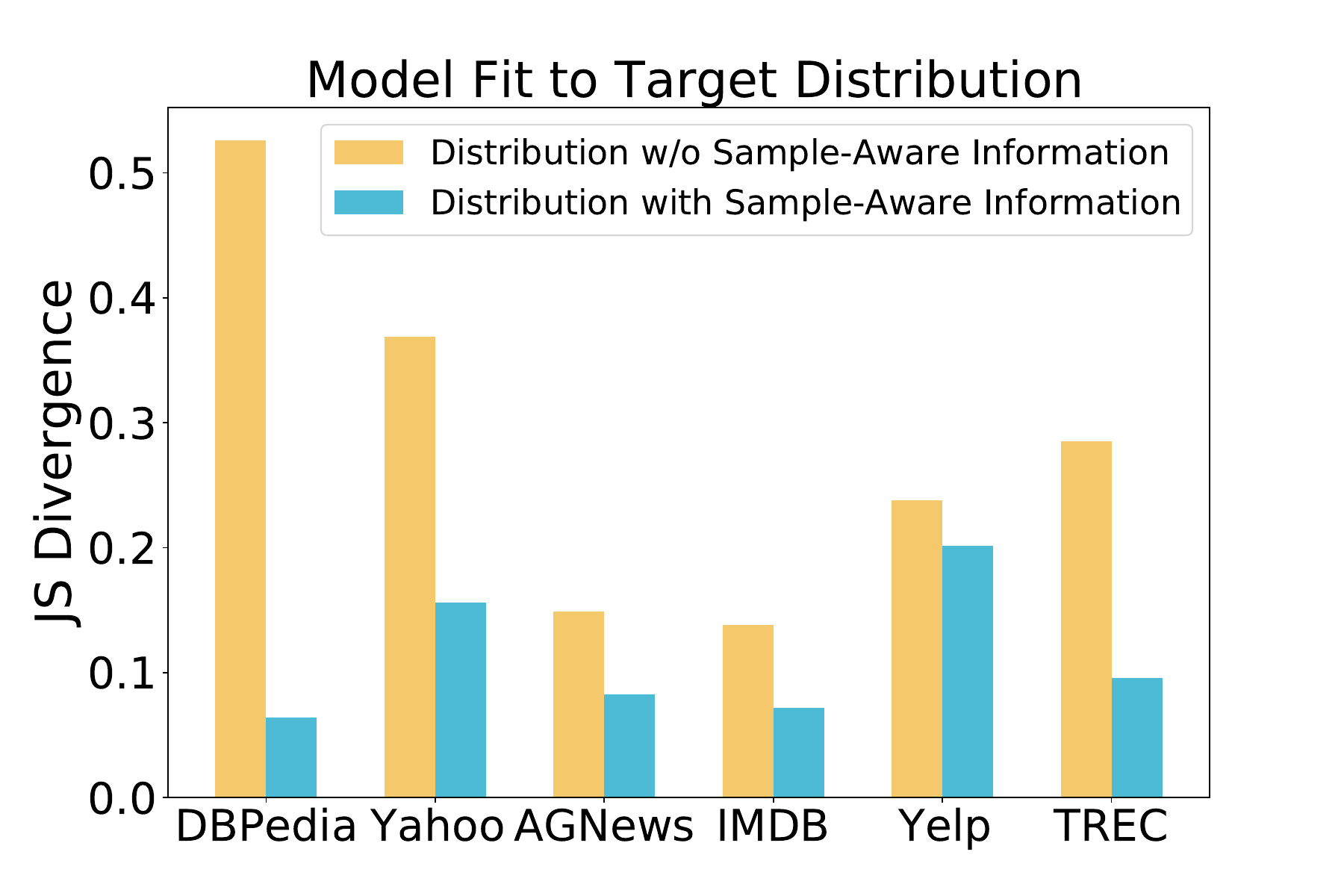}
    \caption{Degree of model fit to the target distribution. The model with unlabeled sample-aware information is consistently closer to the target distribution (with a lower JS divergence value) across six datasets.}
    \label{fig:JS}
\end{figure*}
A distribution alignment experiment was conducted to validate the central motivation of PromptAL. The target distribution was defined as the probability distribution of a fully trained model evaluated on a test dataset. Two model distributions were compared: (1) a model trained exclusively on 32 randomly selected labeled instances and (2) a model trained on the same 32 labeled instances augmented with sample-aware information from unlabeled data. Jensen--Shannon (JS) divergence \cite{61115} was employed to assess the alignment between the trained models and the target distribution, with lower values indicating closer alignment. Figure \ref{fig:JS} shows that across all six datasets, the distribution incorporating sample-aware unlabeled information was significantly closer to the target distribution.

Figure \ref{fig:JS} shows that across all six datasets, the distribution incorporating sample-aware unlabeled information was significantly closer to the target distribution.
Moreover, incorporating sample-aware information had a minimal impact on the Yelp dataset (JS divergence decreased by \textbf{0.036}).
In contrast, the distribution fitting effect was substantially improved in the DBpedia dataset (JS divergence decreased by \textbf{0.462}).
This is attributed to the high semantic similarity among samples in the Yelp dataset, making them difficult to distinguish and failing to provide sufficient diversity for the model.
In contrast, the DBpedia dataset spans more categories with greater sample diversity, rendering the effect of incorporating sample-aware information more significant.
In summary, this experiment validates the beneficial effect of integrating unlabeled sample information to harmonize the current distribution with the target distribution.

\section{Analysis}
In this section, we present a comprehensive analysis of the PromptAL and baseline methods from three perspectives.
(1) In the \textit{Analysis of Selected Samples} module, samples chosen by various AL methods are systematically assessed using multiple metrics.
This evaluation reveals the extent to which PromptAL-selected samples facilitate convergence toward the target distribution.
(2) In the \textit{Computational Cost Analysis} module, we examine the temporal efficiency and  GPU memory cost of PromptAL, identify the primary time-consuming factors, and discuss optimizations that mitigate these costs.
In the \textit{Case Study} module, t-SNE is employed to visualize the embeddings of selected samples in a lower-dimensional space, illustrating how PromptAL effectively enhances sample diversity.
%
%
%
\subsection{Analysis of Selected Samples}
We conducted a thorough analysis of the $|B|$ samples selected via different query strategies, considering category distribution, representativeness, diversity, and uncertainty aspects.
\\
$\diamond$ \textbf{Category distribution}.
Following Patron \cite{yuColdStartDataSelection2023}, we utilize two metrics to evaluate the category distribution of the selected samples: imbalance value (IMB) and label distribution divergence (LDD).
Specifically, let the number of samples selected from each category be denoted as  $n_1, n_2, \dots, n_c.$, where $ \sum_{i=1}^{c} n_i = |B| $.
Below is the IMB formula, where lower values signify a more balanced label distribution.
\begin{equation}
    \text{IMB} = \frac{\max_{i=1,\dots,c}(n_i)}{\min_{i=1,\dots,c}(n_i)}
\end{equation}
The LDD quantifies the discrepancy between the true label distribution and that sampled by AL methods.
Specifically, let $p_i$ denote the frequency of label $i$, and $q_i = n_i / |B|$ represent the frequency of category $i$ within the query set.
The LDD is calculated as follows, with lower values indicating that the sampled distribution more closely aligns with the true label distribution.
\begin{equation}
    \text{LDD} = D_{\text{KL}}(q \parallel p) = - \sum_{i} q_i \log \left( \frac{p_i}{q_i} \right)
\end{equation}
\\
$\diamond$ \textbf{Diversity (Div)}.
To evaluate sample diversity in the feature space, we utilize the \texttt{[CLS]} representations derived from a model trained on the entire training dataset.
The test accuracy results of the fully trained model are presented in Table \ref{fully}.
Following \cite{dor2020active}, we calculate the diversity of $B$ as
$\left( \frac{1}{|U|} \sum_{x_i \in U} \min_{x_j \in B} d\left( \Phi(x_i), \Phi(x_j) \right) \right)^{-1}$. 
Here, $\Phi(x_i)$ denotes the \texttt{[CLS]} representation of example $x_i$, $U$ represents the unlabeled pool, and $d$ signifies the Euclidean distance.
A higher Div value indicates greater diversity.
\\
$\diamond $ \textbf{Representativeness (Rep)}.
The objective of Rep analysis is to determine whether the strategy selects outliers that are unrepresentative of the overall data distribution. We used the KNN density metric proposed by \cite{zhu2008active}, which defines the Rep of a sample as one over the average Euclidean distance between it and its k most similar samples based on the feature space mentioned in Div.
A higher Rep signifies greater representativeness.
\\
$\diamond $ \textbf{Uncertainty (Unc)}.
Consistent with \cite{yuanColdstartActiveLearning2020}, we leverage the logits from the fully supervised model to measure sample uncertainty.
Specifically, the average predictive entropy quantifies the uncertainty of batch $B$, with higher values reflecting increased uncertainty.
\begin{table}[ht]
   \centering
   \caption{Fully supervised accuracy (in \%) on six datasets}
 \resizebox{0.6\textwidth}{!}{
    \begin{tabular}{cccccc}
        \toprule
       DBpedia & Yahoo& TREC & AGNews & IMDB & YELP \\
        \midrule
        99.18     & 75.64 & 96.20&94.97&93.50 
          & 64.19
        \\
        \bottomrule
    \end{tabular}
    }
 
  \label{fully}
\end{table}
\begin{table}[ht]
    \centering
    \caption{Category distribution, representativeness, diversity, and uncertainty analysis for selected samples across AL methods, averaged for all datasets.}      
    \resizebox{0.6\textwidth}{!}{
    \begin{tabular}{ccccccc}
        \toprule
        Method  &IMB$\downarrow$ & LDD$\downarrow$  &Rep.$\uparrow$ & Div.$\uparrow$ & Unc.$\uparrow$ \\
        \midrule
        Random &3.072 &0.008& 0.222 & 0.171 & 0.304 \\
        Entropy &3.676&0.044 & 0.143 & 0.156 & 0.446 \\
        LC &3.471&0.041& 0.142 & 0.157 & 0.435 \\
        BERT-KM &2.558&0.009 & 0.239 & 0.173 & 0.226 \\
        BADGE &3.214 &0.008& 0.146 & 0.172 & 0.433 \\
        Patron &6.715&0.015& 0.297 & 0.174 & 0.306 \\
        CAL &3.811&0.068 & 0.212 & 0.155 & 0.336\\
         { Core-set} &  {2.787 }&  {0.068} &  { 0.238 }&   {0.164} &   {0.316} \\
           {BALD}&{3.576}&{0.038 }& {0.144} & {0.158} &{ 0.433}\\
        \textbf{PromptAL} &\textbf{2.613}& \textbf{0.016} &\textbf{0.248} & \textbf{0.174} & \textbf{0.480} \\
        \bottomrule
    \end{tabular}
    }
    
    \label{tab:methods_metrics}
\end{table}

As listed in Table \ref{tab:methods_metrics}, PromptAL achieves an optimal balance among category distribution, Rep, uncertainty, and diversity, demonstrating substantial improvements across all metrics.
Regarding the IMB and LDD metrics, uncertainty-based methods show elevated values, indicating a significant imbalance in the selected sample categories.
This imbalance directly impairs their performance in few-shot learning scenarios.
In contrast, PromptAL outperforms these methods in both IMB and LDD metrics, attaining an IMB of \textbf{2.613} and an LDD of \textbf{0.016}.

This improvement can be attributed to the dynamic soft prompt mechanism of PromptAL, which optimizes decision boundaries.
This mechanism prevents the selection of all samples from a single category due to initially overly shifted boundaries, leading to a more balanced sample selection and ultimately enhancing few-shot learning performance.

PromptAL achieves the highest uncertainty (\textbf{0.480}) and diversity (\textbf{0.174}), indicating its effectiveness in identifying and selecting samples that are both uncertain and diverse.
This can be attributed to its ability to integrate uncertainty scores and diversity evaluation concurrently.
This dual consideration ensures that the selected samples are both highly informative and thoroughly represent various regions of the data distribution.

In terms of Rep, PromptAL achieves a significant value (\textbf{0.248}), demonstrating its proficiency in selecting samples that represent entire datasets.
It surpasses the diversity-based method BERT-KM (\textbf{0.239}) and the uncertainty-based method LC (\textbf{0.142}), suggesting a more effective capture of the target distribution while avoiding the selection of outlier points.
We ascribe this performance to the joint scoring strategy, which combines uncertainty sampling with local diversity considerations.
This approach ensures that excessively uncertain outliers are excluded, thereby enhancing Rep.

Entropy and LC might select samples with high uncertainty yet similar characteristics, leading to lower Rep.
BERT-KM only accounts for global diversity, resulting in samples with low information content.
In contrast, the hybrid strategy PromptAL achieves a superior balance compared to methods that focus exclusively on one aspect.

\subsection{{Computational Cost Analysis}}
{ We assessed the computational cost of PromptAL by analyzing two metrics: time efficiency and GPU memory.}
The time efficiency of PromptAL is largely influenced by the KNN computation within the Local Diversity module, which accounts for the majority of the overhead.
To mitigate this, we employ the k-d tree algorithm \cite{kdtree} to optimize the nearest neighbor search, which significantly reduces computational overhead.
Table \ref{tab:time} displays the selection time of PromptAL compared to baseline methods on the TREC dataset when querying 32 samples.
Overall, PromptAL delivers superior computational efficiency, exhibiting significantly reduced time overhead relative to CAL and BADGE while effectively balancing performance and efficiency.
Thus, it emerges as a highly competitive solution in few-shot AL.

{ Importantly, regarding GPU usage, PromptAL consistently maintains a minimal memory footprint. Under the experimental settings detailed in Section 5.1.3, peak GPU memory usage during training and querying reaches 5,242 MB, which is well below the capacity of typical consumer-grade GPUs (e.g., those with 12 GB of memory). This demonstrates that PromptAL does not require high-end hardware and is suitable for deployment in resource-constrained environments.}

\begin{table}[!ht]
    \centering
     \caption{Methodological time consumption.}
    \resizebox{0.35\textwidth}{!}{
    \begin{tabular}{cc}
        \toprule
        \textbf{Method}  & \textbf{Time (seconds)} \\
        \midrule
        Random &0.1s \\
        Entropy &27s \\
        LC &27s \\
        BERT-KM &26s \\
        BADGE &31s \\
        Patron &14s \\
        CAL &38s\\
       {Core-set}  & {15s}\\
    {BALD } & {11s} \\
        \textbf{PromptAL} &\textbf{17s} \\
        \bottomrule
    \end{tabular}
     }
    \label{tab:time}
\end{table}

\subsection{Case Study}
\begin{figure*}[!hbt]
    \centering
    \begin{subfigure}{0.48\textwidth}
        \centering
        \includegraphics[width=\linewidth]{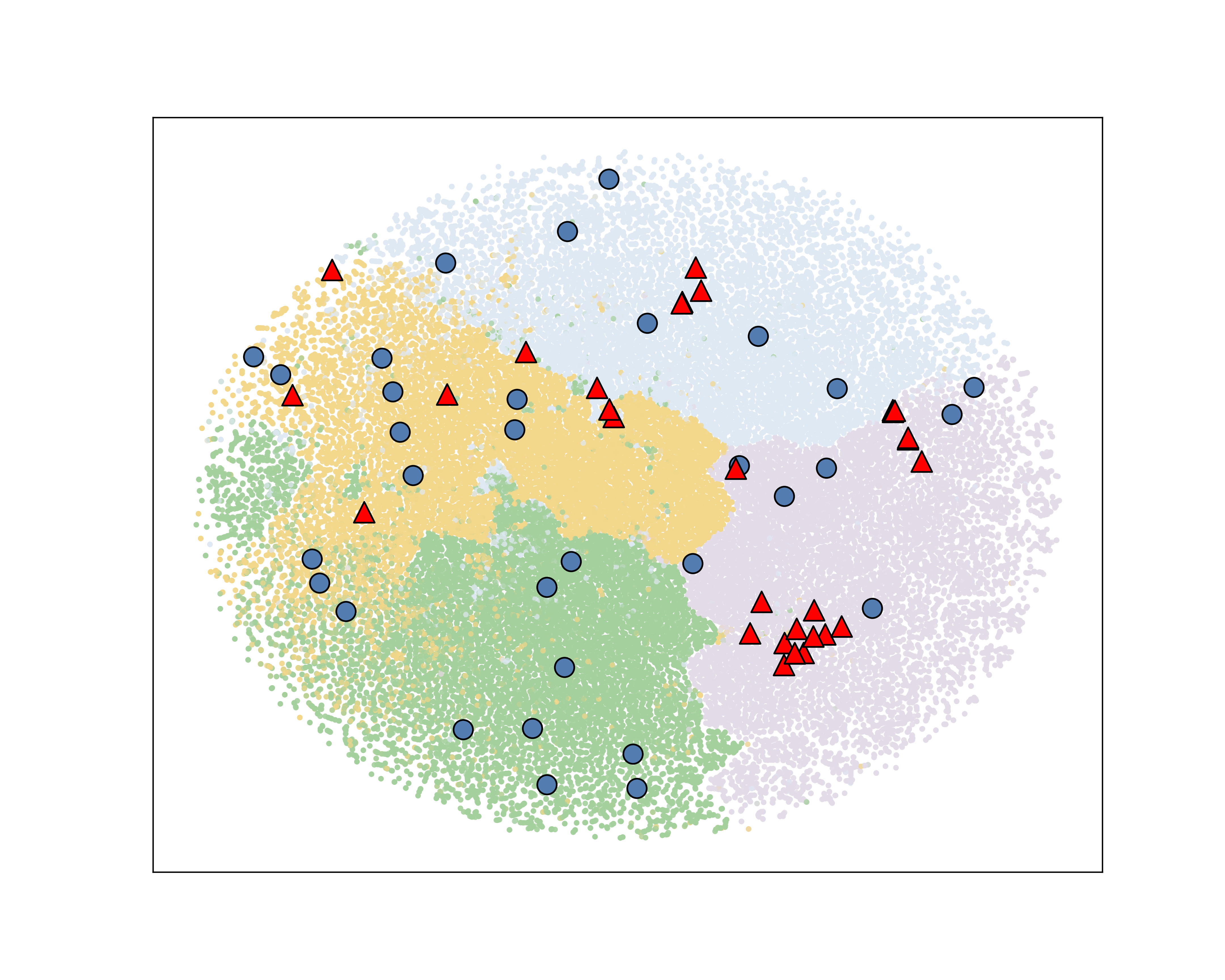}
        \caption{Entropy}
    \end{subfigure}%
    \begin{subfigure}{0.48\textwidth}
        \centering
    \includegraphics[width=\linewidth]{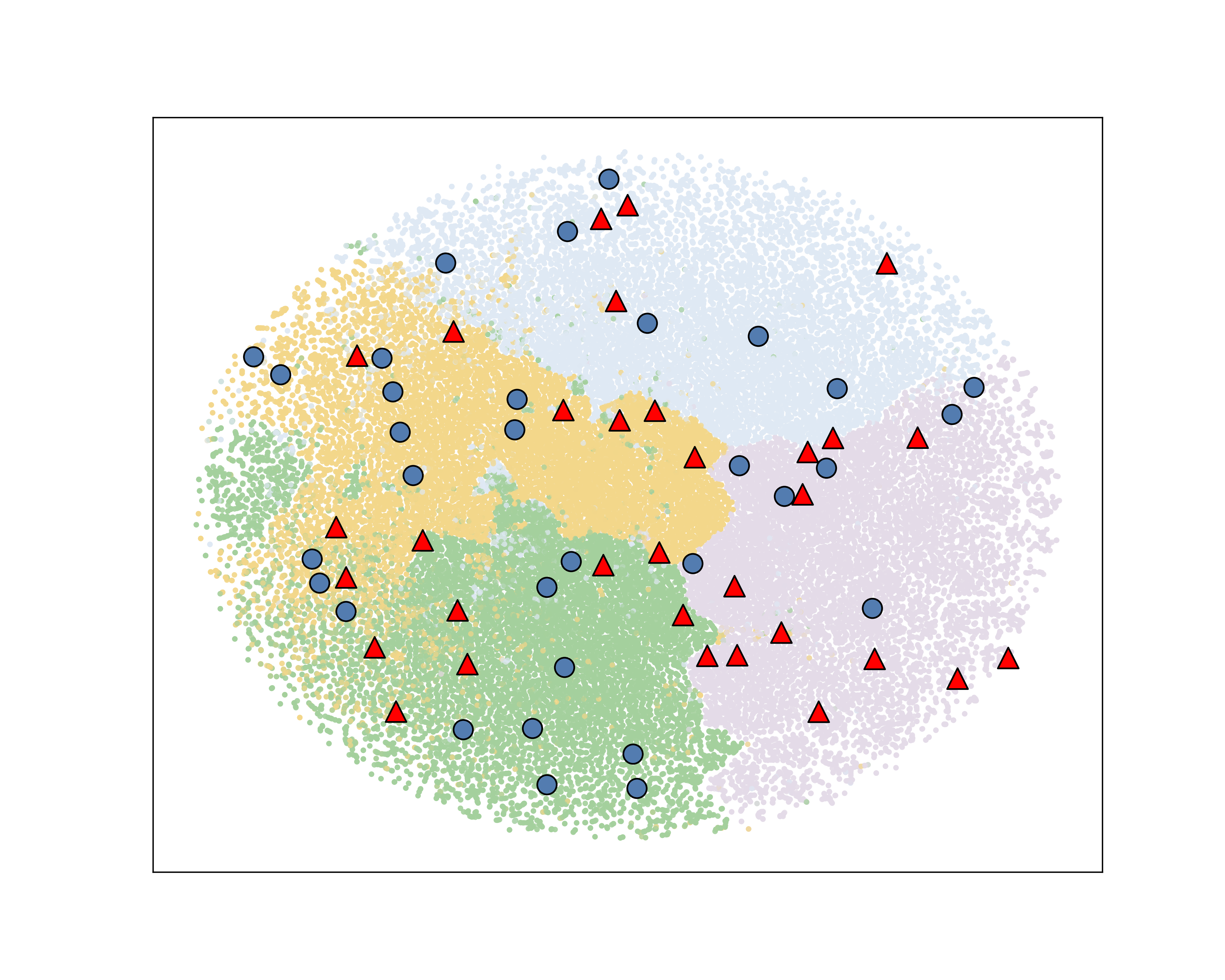}
        \caption{PromptAL}
    \end{subfigure}
    \caption{Samples selected by Entropy and PromptAL after the first AL round on the AGNews dataset. Red triangles denote the selected data points by these methods, the four background colors represent four different categories of unlabeled samples, and blue circles represent the samples from the original training set.}
    \label{case study}
\end{figure*}

We utilized t-SNE \cite{t-SNE} for dimensionality reduction to visualize sentence embeddings within the AGNews unlabeled pool.
Figure \ref{case study} illustrates 32 samples selected by Entropy and PromptAL after the first AL iteration, marked with red triangles.
The four background colors represent four distinct categories of unlabeled samples, while blue circles denote samples from the original training set.
The data points selected by Entropy are densely clustered, indicating substantial similarity and resulting in redundancy.
Moreover, certain samples selected by Entropy are extremely similar to the labeled samples in the training set, which further causes redundancy.
In contrast, PromptAL more effectively captures overall feature diversity, indicating that it successfully enhances both global and local diversity in sample selection.

{
\section{Discussion}
\label{discussion}
This section explores the necessity of PromptAL in the era of large language models (LLMs).
In Section~\ref{sec:model_choice}, we systematically compare RoBERTa-Base, the foundational model of PromptAL, with state-of-the-art LLMs such as Llama 3~\cite{grattafiori2024llama}, Qwen-72B~\cite{qwen2025qwen25technicalreport}, and DeepSeek-V3~\cite{liu2024deepseek}.
Additionally, in Section~\ref{sec:performance_comparison}, we conduct comparative experiments to evaluate the accuracy of PromptAL against zero-shot and few-shot classification performed by leading LLMs across six datasets.
Building on these comparisons, Section~\ref{sec:Advantages} highlights the unique advantages of PromptAL, including computational efficiency and data privacy preservation.
Finally, Section~\ref{sec:necessity} demonstrates the indispensability of PromptAL in specialized domains through application-specific necessity proofs.

\subsection{Comparison of Model Scale and Resource Consumption}
\label{sec:model_choice}
Recent LLMs have shown robust zero-shot and few-shot learning capabilities in NLP, driven by effective prompt designs \cite{sahoo2025systematicsurveypromptengineering}.
In AL scenarios that require frequent parameter updates, compact and high-performance models that are easily deployable, like PromptAL, offer significant practical advantages.
However, owing to the massive number of parameters, open-source LLMs often require a large amount of GPU memory, rendering them impractical for resource-limited environments.
To justify selecting RoBERTa as the backbone model for PromptAL, we perform a systematic comparison presented in Table \ref{Comparison of RoBERTa with Large Language Models} between RoBERTa and LLMs across various dimensions, including parameter count, applicable domains, and computational efficiency.
\begin{table}[ht]
    \centering
    \caption{{Comparison of RoBERTa-base with LLMs. All computational costs are normalized relative to RoBERTa, with its value set as the 1x baseline.}}
    \label{Comparison of RoBERTa with Large Language Models}
    \resizebox{1.0\textwidth}{!}{
    \begin{tabular}{lccccc}
        \toprule
        Model & Activated Parameters & Computational Cost & Fine-tuning Memory Usage & Cross-domain Flexibility & Time per Token \\
        \midrule
        RoBERTa-base & 0.1B & 1$\times$ & $\sim$6--8 GB (full tuning) & Easy \& Fast & 3-5 ms/token \\
        LLaMA3-8B & 8B & $\sim$50$\times$ & $\sim$45--60 GB (LoRA) & High Cost & 50–100 ms/token \\
        Qwen2.5-72B & 72B & $\sim$400$\times$ & $\sim$100--120 GB (LoRA) & High Cost & $\gg$ 50–100 ms/token \\
        DeepSeek-V3-671B & 37B & $\sim$90$\times$ & $\sim$40--60 GB (LoRA) & High Cost & $\gg$ 50–100 ms/token \\
        \bottomrule
    \end{tabular}
    }
\end{table}


$\diamond$ \textbf{Parameter-scale Comparison.} 
The RoBERTa-base model comprises approximately 0.1 billion parameters, whereas LLaMA3-8B and Qwen2.5-72B feature 8 billion and 72 billion parameters, respectively, reflecting a parameter difference of approximately 80 to 720 times.
This disparity leads to significant variations in memory consumption during both fine-tuning and inference.
Even with parameter-efficient tuning methods like LoRA~\cite{hu2022lora}, LLMs necessitate several times more GPU memory compared to models such as RoBERTa.

$\diamond$ \textbf{Quantified Computational Efficiency.} 
Additionally, RoBERTa exhibits superior computational efficiency during inference relative to autoregressive LLMs, as computational time is primarily influenced by the number of parameters activated during inference.
For example, while models with comparable AL performance, such as LLaMA3-8B, contain 8 billion parameters, RoBERTa operates with only 0.1 billion.
Comparative experiments were conducted using the vLLM framework \cite{vllm} and an NVIDIA A100 GPU (80 GB).
The results indicate that RoBERTa achieves an inference latency of merely 3–5 milliseconds per token, whereas LLaMA3-8B requires 50–100 milliseconds per token.
Qwen2.5-72B and DeepSeek-V3 require more GPU resources and exhibit even longer inference times.
This indicates significantly higher computational costs for the larger models compared to RoBERTa.

$\diamond$ \textbf{Domain-adaptation Empirical Evidence.}
Although LLMs exhibit impressive zero-shot and few-shot capabilities, studies reveal that their performance in highly specialized domains often lags without domain-specific fine-tuning~\cite{bang2023multitask,short1}.
In contrast, smaller models can be effectively fine-tuned on target domain data, delivering strong adaptability and reliable performance across various professional fields~\cite{chalkidis2020legal,lee2020biobert}.
This adaptability is crucial in AL scenarios, where models undergo iterative training through multiple rounds as new labeled samples are incorporated.
In such contexts, smaller models like RoBERTa offer significant advantages in fine-tuning efficiency, resource consumption, and deployment flexibility.

Consequently, PromptAL selected RoBERTa-base for its balanced parameter scale, domain versatility, and ease of deployment, which align with the demands of AL for rapid model iteration and provide clear engineering benefits in resource-constrained environments.
\subsection{Performance Comparison with Large Language Models}
\label{sec:performance_comparison}
We conducted a systematic comparative analysis of PromptAL's performance against leading LLMs—including GPT-4o~\cite{openai2024hellogpt4o}, Claude-3~\cite{Claude}, and DeepSeek-R1~\cite{guo2025deepseek}—under zero-shot and few-shot settings.
The experimental configurations for all LLMs are presented in Table~\ref{set_llms}, and experiments were performed on six datasets, detailed in Table~\ref{tab:dataset}.
In the 5-shot setting, five samples are randomly selected from the training set as in-context demonstrations and included in the prompt for each query. The accuracy results are presented in Table~\ref{acc_llms}.
\begin{table}[h]
\centering
\caption{{Experimental settings for LLM evaluation.}}
\label{set_llms}
\resizebox{0.6\textwidth}{!}{
\begin{tabular}{ll}
\toprule
\textbf{Setting} & \textbf{Value} \\
\midrule
GPT-4o Version & \texttt{chatgpt-4o-latest} \\
Claude-3 Version & \texttt{claude-3-7-sonnet-20250219} \\
DeepSeek-R1 Version & \texttt{deepseek-r1-250528} \\
Temperature & 0.5 \\
Top-p  & 0.7 \\
Max Input Tokens & 2048 \\
Evaluation Metric & Accuracy \\
\bottomrule
\end{tabular}}
\end{table}
\begin{table}[htbp]
\centering
\caption{ {
Accuracy (in \%) and API costs of PromptAL and LLMs in few-shot and zero-shot settings across six datasets. API costs are determined by the number of input and output tokens and reflect pricing as of June 29, 2025. The API costs are reported in US dollars per million input tokens, with “/MTok" denoting the unit million tokens. Note that the cost of output tokens is typically higher than that of input tokens, which further increases the total inference expense.
}}
\label{acc_llms}
\resizebox{0.9\textwidth}{!}{
\begin{tabular}{llccccccc}
\toprule
Model & Setting & IMDB & AGNews & TREC & DBPedia & Yahoo & Yelp& API Cost\\
\midrule
Claude-3      & Zero-shot  & 93.67 &  70.17&55.20  & 93.50 & 67.10 & 67.50&\$3.00  \\
              & 5-shot     &93.00  &87.35  &  57.80& 95.50 &69.37  &  \textbf{69.63}&/MTok\\
\midrule
DeepSeek-R1   & Zero-shot  & 94.44 &82.60 & 58.00 &  89.75& 64.75 & 57.63&\$0.58 \\
              & 5-shot     & \textbf{96.50} & 87.04  &59.80  &93.00  & 68.00 & 64.38&/MTok \\
\midrule
GPT-4o        & Zero-shot  & 93.10 & 87.50 & 59.40 & 93.25 & 68.33 &63.50&\$5.00  \\
              & 5-shot     & 95.63 & 86.16 & 61.80 & 97.40 & \textbf{69.84} &  68.12&/MTok\\
\midrule
PromptAL       & Round 5   & 89.39 &89.17  & 91.60 & 98.16 & 66.01 & 51.56&\$0.00   \\
               & Round 10  &89.68  &\textbf{89.88}  & \textbf{94.60} &  \textbf{98.44}& 67.34 & 52.95&/MTok\\
\bottomrule
\end{tabular}}
\end{table}

PromptAL outperforms LLMs in question answering, knowledge-base classification, and news topic categorization tasks (e.g., TREC, DBPedia, and AGNews).
Conversely, LLMs excel in sentiment classification tasks such as IMDB and Yelp.
In terms of cost, PromptAL supports low-cost local deployment and is substantially more economical than online services that charge per token, including Claude-3, DeepSeek-R1, and GPT-4o.
Overall, PromptAL achieves comparable or superior accuracy to these LLMs across most datasets while maintaining substantially lower costs.
\subsection{Advantages of PromptAL Compared to LLM-based Paradigms}\label{sec:Advantages}
Additionally, PromptAL offers significant advantages over LLMs in computational efficiency and data privacy preservation.
\\
$\diamond$ \textbf{Computational Efficiency.}
PromptAL exhibits substantial advantages in computational efficiency compared to LLMs.
As demonstrated in Section \ref{sec:model_choice}, while open-source LLMs support deployment and fine-tuning, both the training and inference stages incur considerable computational resources and memory consumption, posing challenges for deployment in resource-limited settings.
Conversely, PromptAL utilizes a model with fewer parameters, significantly lowering computational costs.

Additionally, Section \ref{sec:performance_comparison} highlights that commercial LLMs are typically accessed via remote APIs for inference, which introduces network latency.
When handling large-scale datasets, non-computational time expenses—particularly data transmission and network latency—can greatly surpass actual inference time, becoming the main efficiency bottleneck.
In contrast, PromptAL employs small models to perform all inference locally, thereby eliminating network-related delays and significantly enhancing efficiency.
\\
$\diamond$ \textbf{Data Privacy Preservation.}
In highly sensitive sectors such as healthcare, government affairs, and finance, data often involve user privacy and legal compliance requirements that strictly prohibit cross-domain transmission.
These constraints make using online APIs like GPT-4 and Claude challenging.

Developers cannot control data flow, increasing the risk of data leakage, and interacting with external APIs may violate privacy regulations such as the U.S. Health Insurance Portability and Accountability Act (HIPAA) and the European Union’s General Data Protection Regulation (GDPR).
Instead, PromptAL is an AL framework based on RoBERTa, enabling high-quality sample selection and iterative model training within local environments.
This framework ensures that raw data remain within the domain and inherently complies with scenarios demanding rigorous data security.

Overall, in the era of LLMs, PromptAL consistently showcases notable benefits in computational efficiency and data privacy.
\subsection{Domain-Specific Necessity for PromptAL}\label{sec:necessity}
While LLMs demonstrate remarkable few-shot and zero-shot abilities, certain critical applications still deem PromptAL essential.
In highly specialized and sensitive fields, labeled data is often scarce.
For instance, dark web data are extremely sensitive and difficult to share publicly, and rare disease samples usually require expert medical annotation.
Consequently, due to the acute shortage of labeled samples in the initial phases, AL becomes a viable strategy.

On one hand, fine-tuning LLMs with domain-specific labeled data for AL entails significant computational expenses, making this method impractical for resource-constrained organizations.

On the other hand, using AL techniques other than PromptAL with limited labeled training data may alter decision boundaries, leading to skewed sample selection by conventional AL methods and ultimately degrading the model's overall performance.
In such scenarios, PromptAL remains an indispensable AL approach.
By integrating sample-aware dynamic soft prompts, PromptAL effectively directs the model to adjust decision boundaries.
It simultaneously assesses unlabeled samples based on uncertainty and diversity, facilitating the selection of high-quality data to enhance the training set.
This approach enables swift performance gains under low-resource and privacy-restricted conditions.

In conclusion, Section \ref{discussion} demonstrates that PromptAL remains a practical and indispensable AL framework in the era of LLMs.
PromptAL effectively enhances performance in few-shot scenarios, ensures data privacy, and exhibits adaptability within specialized domains.
}
\section{Conclusions}
In this study, we introduce a novel AL framework, PromptAL, which integrates sample-aware dynamic soft prompts.
These prompts leverage information from unlabeled samples to better align the current distribution with the target distribution, thereby optimizing the decision boundary.
Consequently, PromptAL selects samples that significantly accelerate convergence toward the target distribution.
Comprehensive experiments across six in-domain and three OOD natural language datasets demonstrate that PromptAL outperforms seven baseline methods in few-shot learning contexts.
Moreover, analysis of the selected samples reveals that PromptAL achieves an optimal balance among category distribution, representativeness, uncertainty, and diversity metrics.
Additionally, we conduct extensive ablation studies to systematically evaluate the individual contributions of each PromptAL component, thereby substantiating the methodological robustness and effectiveness of our proposed framework.
{Finally, we systematically analyze PromptAL's advantages over LLMs from multiple perspectives---including computational efficiency, privacy protection, and scenario adaptability---thereby further highlighting its necessity and broad application value in the era of LLMs.}






\bibliographystyle{elsarticle-num}  
\bibliography{PromptAL}    




\end{document}